%% file: main.tex
\documentclass[10pt,twocolumn,letterpaper]{article}

\usepackage[pagenumbers]{cvpr} 

\usepackage{nicefrac}
\usepackage{amsmath} 


\input{notation}

\definecolor{cvprblue}{rgb}{0.21,0.49,0.74}
\usepackage[pagebackref,breaklinks,colorlinks,allcolors=cvprblue]{hyperref}

\def\paperID{10931} 
\def\confName{CVPR}
\def\confYear{2025}

\title{Dynamic Neural Surfaces for Elastic 4D Shape Representation and Analysis}

\author{
Awais Nizamani$^{1*}$ \quad
Hamid Laga$^{1}$ \quad
Guanjin Wang$^{1}$ \quad
Farid Boussaid$^{2}$ \\
Mohammed Bennamoun$^{2}$ \quad
Anuj Srivastava$^{3}$ \\
$^1$Murdoch University \quad
$^2$The University of Western Australia \quad
$^3$Florida State University
}

\begin{document}
\maketitle



\begin{abstract}
We propose a novel framework for the statistical analysis of genus-zero 4D surfaces, \ie 3D surfaces that deform and evolve over time. This problem is particularly challenging due to the arbitrary parameterizations of these surfaces and their varying deformation speeds, necessitating effective spatiotemporal registration. Traditionally, 4D surfaces are discretized, in space and time, before computing their spatiotemporal registrations, geodesics, and statistics. However, this approach may result in suboptimal solutions and, as we demonstrate in this paper, is not necessary. In contrast, we treat 4D surfaces as continuous functions in both space and time. We introduce  Dynamic Spherical Neural Surfaces (D-SNS), an efficient smooth and continuous spatiotemporal representation for  genus-0 4D  surfaces. We then demonstrate how to perform core  4D shape analysis tasks such as spatiotemporal registration, geodesics computation, and mean 4D shape estimation, directly on these continuous representations without upfront discretization and meshing. By integrating neural representations with classical Riemannian geometry and statistical shape analysis techniques, we provide the building blocks for enabling full functional shape analysis. We demonstrate the efficiency of the framework on 4D human and face datasets.  The source code and additional results are available at \url{https://4d-dsns.github.io/DSNS/}.
\end{abstract}

\section{Introduction}
\label{sec:intro}



Statistical 3D shape analysis focuses on quantifying shape similarities and differences between 3D objects. It also aims at modelling, using probability distributions,  the shape variability within and across object classes.  4D shape statistics, on the other hand, adds the temporal dimension. It aims to discover typical shape deformation patterns in a class of objects, and statistically model the variability of these deformations within and across classes. This is a very challenging problem since 3D shapes come with arbitrary discretization and thus need to be spatially registered onto each other. 4D shapes add another level of complexity since objects  deform and grow at different rates. Thus, they need to be temporally aligned onto each other. Traditionally, 3D and 4D shapes are represented as discrete point clouds or triangular meshes. The spatiotemporal registration problem is then reduced to that of matching landmarks across shapes. Shape statistics such as means and modes of variation are then computed directly on these registered landmarks. This, however, leads to solutions that depend on the quality and resolution of the discretization. 

In this paper, we treat 4D shapes as continuous functions in both space and time and develop a novel statistical analysis framework that operates directly on these continuous representations, without upfront discretization. This includes performing spatiotemporal registration, computing geodesics, and estimating the mean 4D shape of a set of 4D surfaces. We focus on genus-0 surfaces, which are abundant in nature, \eg human bodies and body parts. We propose a novel surface-based neural representation, hereinafter referred to as Dynamic Spherical Neural Surface (D-SNS), that treats a 4D surface as a continuous mapping from a spherical domain and time to  3D. This mapping is then parameterized using Multi-Layer Perceptrons (MLPs). This will allow us to formulate the spatial registration problem as that of finding the optimal spatial reparameterizations of the neural functions. Similarly, temporal registration can be formulated as the problem of optimal temporal reparameterization of these neural functions. This can be efficiently implemented using an MLP and optimized using an elastic Riemannian metric that quantifies physical deformations of surfaces, \ie bending and stretching. Additionally, we propose a framework that  co-registers a set of 4D surfaces, represented with their D-SNSs, and simultaneously computes their 4D mean as a D-SNS.  The main contributions of this paper can be summarized as follows; 
\begin{itemize}
    \item We introduce a new continuous representation for genus-0 4D surfaces, parameterized using Multi-Layer Perceptrons (MLPs). This framework treats both 3D and 4D shapes as functions, enabling functional shape analysis. We also demonstrate how this representation can be efficiently learned from discrete input 4D surfaces, with performance validated on datasets such as 4D facial shapes~\cite{VOCA2019,COMA:ECCV18} and 4D human bodies~\cite{CAPE:CVPR:20,dfaust:CVPR:2017}.

    \item We reformulate  the spatiotemporal registration problem as one of optimally reparameterizing, in space and time,  the neural representation. We measure the optimality using a physically-motivated Riemannian elastic metric. 
    
    \item We show that by mapping our representation to the space of Square Root Normal Fields (SRNF)~\cite{laga2017numerical,jermyn2017elastic} and Square Root Velocity Fields (SRVF)~\cite{su2014stat, su2017squarerootvelocityframework}, the complex Riemannian elastic metric simplifies to an $\ltwo$ metric, which facilitates various downstream applications. 
    

    \item We develop a comprehensive framework for performing spatiotemporal registration and computing geodesics and 4D means directly from the neural surface representation without discretization. 
    
\end{itemize}

\noi The remainder of the paper is organized as follows; Section~\ref{sec:relatedwork} discusses the related work. Section~\ref{sec:approach} describes the proposed approach. Section~\ref{sec:results} presents the results of the proposed approach, analyzes its performance, and compares it to the state-of-the-art. Section~\ref{sec:conclusion} concludes the paper.
 
\section{Related Works}
\label{sec:relatedwork}

\subsection{Statistical 3D and 4D shape analysis} 
Early statistical 3D shape analysis methods, based on morphable models~\cite{Cootes1995ASM,blanz1999morphable,allen2003space}, treat 3D shapes as a set of discrete landmarks in correspondence and use the $\ltwo$ metric  for computing geodesics and Principal Component Analysis (PCA) for their statistical analysis. These methods, however, assume that the 3D shapes are elements of an Euclidean shape space. As a result, they cannot handle 3D shapes that undergo large elastic deformations such as bending and stretching. SMPL model and its variants~\cite{loper2015smpl, zuffi2019three, Pavlakos_2019_CVPR, li2017learning} address 
this issue by explicitly modelling articulated motion using skeleton joints. SMPL-based methods, however, are class-specific and thus, they do not allow cross-category analysis.

A common property of these methods is that they assume that correspondences are given, \ie either they are manually specified or computed in a processing step using other techniques, \eg~\cite{bronstein2006multi, Ovsjanikov2012Functional, Ovsjanikov2016Functional}. Jermyn \etal~\cite{jermyn2017elastic} and later Laga \etal~\cite{laga2017numerical},   treat 3D shapes as elements of a Riemannian shape space  equipped with an  elastic metric that measures bending and stretching. This way, registration,  geodesics, and summary statistics computation can be jointly formulated as an optimization problem under the same elastic metric. More importantly, they showed that by further mapping the shapes to the space of Square Root Normal Fields (SRNFs), the complex elastic metric becomes an $\ltwo$ metric, thus, significantly simplifying  downstream applications~\cite{laga2017numerical,laga2018survey}.
The SRNF representation has been used for the analysis of genus-0 surfaces that undergo nonrigid motion. They have also been extended to the analysis of 4D surfaces~\cite{laga20234datlas} by treating them as time-parameterized trajectories in the SRNF space, which has an $\ltwo$ structure. Other methods represent  4D surfaces as the flow of deformation of the 3D volume and compute geodesics on a Riemannian manifold~\cite{beg2005computing,deba2019clustering,bone2020spatiotemp}. Unfortunately, computing deformations on 3D volumes is computationally expensive. 

All these methods  treat 4D shapes as discrete, in space and time, signals. In this paper, we consider, for the first time, 4D shapes as continuous functions and derive a comprehensive framework that operates directly on these functions. Discretization is only required for visualization.

\subsection{Neural representations} 

Implicit neural representations such as DeepSDF~\cite{park2019deepsdf}, Neural Radiance Fields (NeRF)~\cite{mildenhall2020nerf}, and Neural Implicit Surfaces (NeuS)~\cite{wang2021neus}, leverage neural networks to represent the geometry and appearance of 3D shapes as continuous functions. They have been extensively used for 3D reconstruction  but also have the potential to be used for 3D shape analysis tasks. In particular, some papers used neural networks to overfit individual SDFs~\cite{davies2021on}. This leads to a continuous and diffentiable representation of individual shapes. While SDFs represent 3D shapes of arbitrary topologies, they are volumetric representations and thus expensive to evaluate since, in general, the surface of interest occupies just a tiny proportion of the 3D space. As such, several papers have explored neural representations of explicit surfaces. In particular, AtlasNet~\cite{groueix2018papier}   models surfaces as a collection of parametric patches. Each patch is treated as a continuous mapping from a 2D domain to 3D and thus can be parameterized using an MLP. Morreale \etal~\cite{morreale2021neural} introduced Neural Surface Maps, which represent a surface as a mapping of a unit disk to 3D and then overfits to it an MLP. This representation, which  enables surface-to-surface mapping,  has been  extended to computing semantic maps between genus-0 surfaces~\cite{morreale2024neural} and for 3D reconstruction~\cite{zhang2021ners}.  Williamson \etal~\cite{williamson2024sns} showed how to compute differential properties of surfaces and perform geometry processing tasks on  spherical neural representations without upfront discretization.

Our approach follows~\cite{morreale2021neural,achlioptas2018learning}. However, our \textbf{first} key novelty is the generalization of Spherical Neural Surfaces to dynamic surfaces,  \ie 3D surfaces that deform over time, and show how to learn it from discrete 4D surfaces.  We refer to this representation as Dynamic Spherical Neural Surfaces (D-SNS). Our \textbf{second} key novelty is that we use this representation to develop a comprehensive framework for the statistical analysis of 4D surfaces represented with their D-SNS. Traditionally, statistical  shape analysis is performed on discretized surfaces. We show that this is not necessary and propose a framework that enables for the first time to perform functional shape analysis.

\section{Method}
\label{sec:approach}

We introduce a novel neural representation, termed Dynamic Spherical Neural Surfaces, to represent  4D (\ie 3D $+$ time) surfaces (Section~\ref{sec:neural_surfaes}). The representation is continuous in space and time and thus enables us to treat 4D surfaces as continuous functions.
To enable the statistical analysis of such 4D surfaces, we introduce a Riemannian shape space, equipped with a Riemannian elastic metric that measures bending and stretching. 3D surfaces can be seen as points in the Riemannian shape space while 4D neural surfaces can be seen as trajectories in that space. The shape space, however, is of infinite dimension and has a nonlinear structure, which makes downstream analysis tasks complex and computationally expensive. Thus, we further map 3D  surfaces to the space of Square-Root Normal Fields (SRNFs)~\cite{laga2017numerical}, which has an $\ltwo$ structure. More importantly, the $\ltwo$ metric in the space of SRNFs is equivalent to the partial elastic metric in the original space. This way, 4D neural surfaces become trajectories in the SRNF space, which is Euclidean. Thus, spatiotemporal registration (Section~\ref{sec:registration}), geodesics between neural 4D surfaces (Section~\ref{sec:geodesics}), and summary shape statistics of neural 4D surfaces (Section~\ref{sec:statistics}) can be efficiently computed in the SRNF space and mapped back to the original space for visualization. Importantly, we show that these quantities can be directly computed on the neural representation without upfront discretization.

\subsection{Dynamic Spherical Neural Surfaces (D-SNS)}
\label{sec:neural_surfaes}
We treat a manifold surface, after normalization for translation and scale, as a function $\surface: \domain \to [-1, 1]^3$, where $\domain \subset \rthree$ is a parameterization domain. Let $\preshapesf$ be the pre-shape space of such functions. When dealing with closed genus-0 surfaces, $\domain$ can be defined as a unit sphere $\stwo$. Dynamic surfaces, \ie surfaces that deform over time can be represented in the same way by adding the temporal dimension, \ie $\neuralrepresentationswithoutparameters: \domain \times [0, 1] \to [-1, 1]^3$ where $[0, 1] $ is the normalized time domain. Traditionally, $\surface$ and $\neuralrepresentationswithoutparameters$ are discretized both in space, and space and time, respectively~\cite{laga2017numerical,laga20234datlas}. Thus, the performance of the subsequent analysis tasks depends on the quality of the discretization. In this paper, we exploit the power of neural networks as universal approximators to represent  $\neuralrepresentationswithoutparameters$ as a continuous function in space and time. We refer to this novel neural representation as \emph{Dynamic Spherical Neural Surfaces (D-SNS)}.

\begin{figure}[t]
  \centering
   \includegraphics[trim=0 180 0 50, width=1\linewidth]{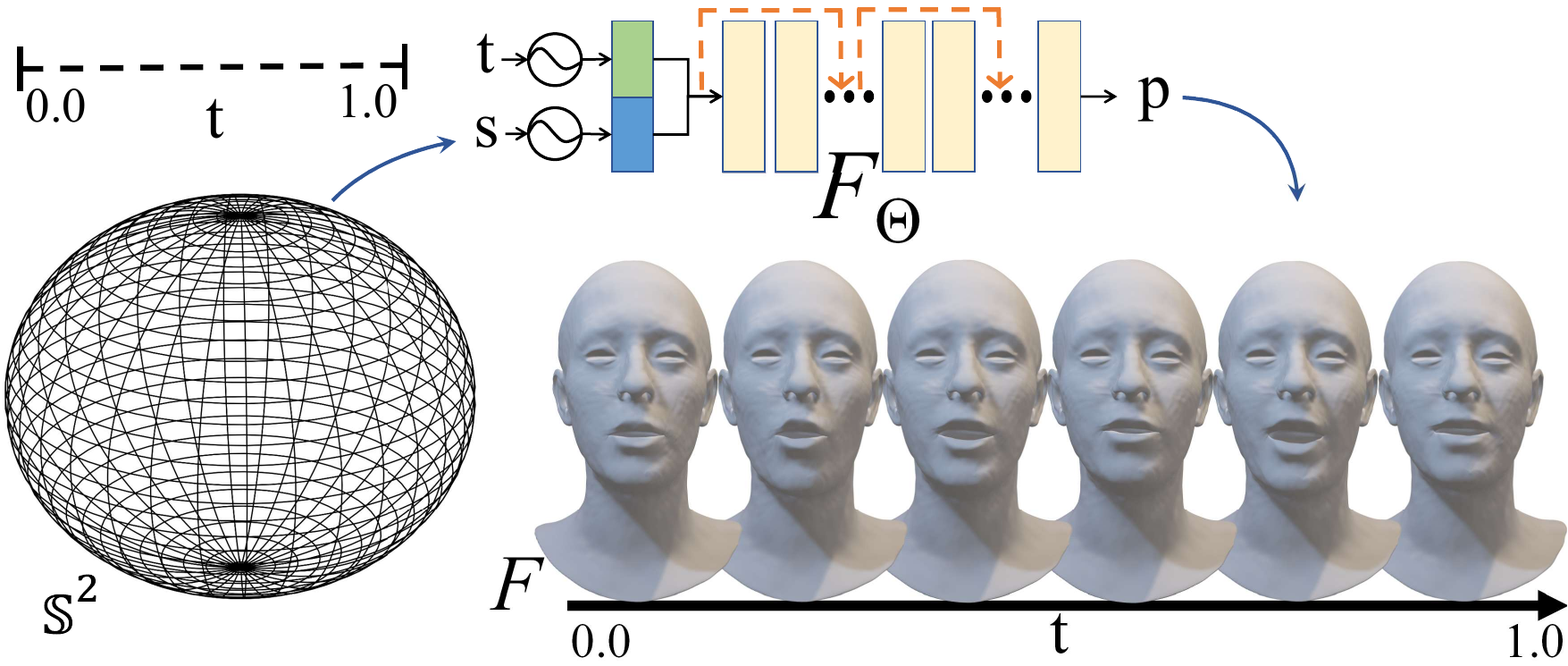}\\
   4D surface $\spatiotemporalsurface: \stwo \times [0,1] \to [-1,1]^3$.

   \caption{\textbf{Dynamic Spherical Neural Surface. }Given a discrete 4D surface of meshes parameterized by a unit sphere $\stwo$ and time $t$ - we overfit an MLP $\neuralrepresentations$ to create a continuous 4D surface $\spatiotemporalsurface$ by minimizing the Mean Square Error between the ground truth and predicted surface points.}
   \label{fig:neuralrepresentation}
\end{figure}

\subsubsection{Representation} 

Given a discrete 4D surface where each time instance is a genus-0 triangular mesh, we first spherically parameterize each individual mesh using the approach of Praun and Hoppe~\cite{praun2003spherical,kurtek2013landmark} and put them in correspondence using the approach of Laga \etal~\cite{laga2017numerical}.  To learn a continuous representation, we overfit a Multi-Layer Perceptron (MLP) $\neuralrepresentations$ to this representation. The MLP maps a  $\pointonsphere \in \stwo$ and a time  $\thetime \in [0, 1]$ to a 3D surface point $\point\in [-1, 1]^3$:
\begin{equation}
  \neuralrepresentations: \stwo \times [0,1] \to [-1, 1]^3; \quad
  \pointonsphere, \thetime \mapsto \neuralrepresentations(\pointonsphere,\thetime) = \pointonsurface.
  \label{eq:D-SNS}
\end{equation}

\noi Here, $\pointonsphere = (\spherecoordtheta,\spherecoordphi)$ are  the spherical coordinates and $\parameters$  the network parameters. We overfit the MLP  to a discrete 4D surface by minimizing the Mean Square Error between the ground truth point $\singlepointonsurface$ and the predicted point $\predictedsinglepointonsurface$:
\begin{equation}
  \neuralloss = \frac{1}{K}\frac{1}{N}\sum_{j=1}^K \sum_{i=1}^N \parallel \neuralrepresentations( \singlepointonsphere,t_j) - 
  p_{ij} \parallel^2. 
  \label{eq:D-SNSloss}
\end{equation}

\noi Here, $N$ is the number of sample points per surface and  $K$ is the number of temporal samples used for training.

One advantage of the proposed neural representation is that differential properties of 3D and 4D surfaces can be computed directly in the continuous domain, without discretization. In particular, the normal field of a 3D surface $\neuralrepresentations(\cdot, \thetime)$, which is essential for the spatiotemporal registration as will be seen in Section~\ref{sec:registration},  is given by:
\begin{equation}
    \nabla \neuralrepresentations(\cdot, \thetime) = \left(\frac{\partial \neuralrepresentations (\cdot, \thetime)}{\partial \spherecoordtheta}, \frac{1}{\sin\spherecoordtheta} \frac{\partial \neuralrepresentations (\cdot, \thetime) }{\partial \spherecoordphi}\right)^\top,
    \label{eq:normal_field}
\end{equation}

\noi where $\spherecoordtheta \in [0, \pi]$ and $\spherecoordphi \in [0, 2\pi)$. Similarly, the tangent field to the 4D surface can be computed by taking the derivative with respect to time, \ie $\nicefrac{\partial \neuralrepresentations}{\partial \thetime}$. These differential quantities can be automatically computed using the automatic differentiation functionality built into modern machine learning setups.

\subsubsection{Network architecture}
\label{sec:D-SNSnetwork}

We use an MLP consisting of six residual blocks. Each block has two layers, with 1024 nodes each. The use of residual connections significantly enhances the network's ability to represent  coarse and fine geometric details. It also significantly improves convergence time and reduces the risk of converging to local minima.  Inspired by Space-time NeRF~\cite{xian2021space}, we apply positional encoding to the spatial domain $\stwo$ and to the temporal domain $[0, 1]$. To avoid sharp edges typically associated with the ReLu activation function, we use the SoftPlus activation function~\cite{zheng2015improving}, which is smooth and continuous. This way, the neural network represents 4D surfaces with very high accuracy.



\subsection{Spatiotemporal registration using D-SNS}
\label{sec:registration}

\begin{figure}[t]
  \centering
   \includegraphics[trim=270 200 260 200, clip, width=1\linewidth]{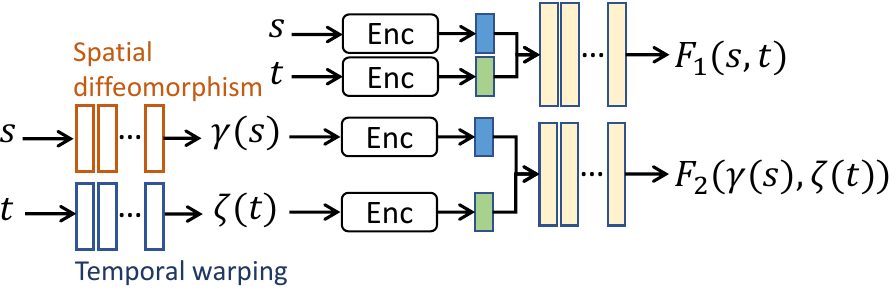}

   \caption{Illustration of the spatiotemporal registration framework proposed in this paper. "Enc" refers to positional encoding.}
   \label{fig:spatialtemporalregistration}
\end{figure}

In this section, we show how to spatiotemporally register two 4D surfaces $\spatiotemporalsurface_1$ and $\spatiotemporalsurface_2$ directly in the continuous domain using their D-SNS representation.  We formulate the spatiotemporal registration problem as that of finding a rotation $\rotation \in \rotations$, a spatial diffeomorphism $ \diffeo: \stwo \to \stwo$, and a temporal diffeomorphism $\timewarp: [0, 1] \to [0, 1]$ that bring the two 4D surfaces as close as possible to each other. This is an optimization problem of the form:
\begin{equation}
    (\rotation^*, \diffeo^*, \timewarp^*) = \underset{\rotation, \diffeo, \timewarp}{\argmin} \int_0^1\loss\left(\spatiotemporalsurface_1(\thetime),  \rotation(\spatiotemporalsurface_2 \circ \timewarp)(\thetime) \circ  \diffeo\right)\mathrm{d}\thetime.
    \label{eq:spatiotemporal_registration_loss}
\end{equation}

\noi Here, $\loss$ is a measure of closeness. In this paper, we treat $\diffeo$ and $\timewarp$ as continuous functions; see Figure~\ref{fig:spatialtemporalregistration}. During training, we freeze the D-SNS networks that represent $\spatiotemporalsurface_1$ and $\spatiotemporalsurface_2$, and only optimize the parameters of $\diffeo$ and $\timewarp$, using the metric $\loss$ as the loss function. This is, however, not straightforward and can be very time-consuming since the optimization needs to be performed over the entire temporal sequence and using a metric $\loss$ that is not linear. Thus, we first proceed with spatial registration (Section~\ref{sec:spatial_registration}) followed by temporal registration (Section~\ref{sec:temporal_registration}).

\subsubsection{Spatial registration} 
\label{sec:spatial_registration}
For simplicity of notation, let  $\surface_1 = \spatiotemporalsurface_1(\thetime)$ and $\surface_2 = \spatiotemporalsurface_2(\thetime)$.
To spatially register $\surface_2$ onto $\surfaceone$, we need to find a spatial diffeomorphism $\diffeo^*: \stwo \to \stwo$ and a rotation $\rotation^* \in \rotations$ that bring $\surfacetwo$ as close as possible to $\surface_1$, \ie:
\begin{equation}
  (\rotation^*,\diffeo^*) = \underset{\rotation,\diffeo}{\argmin}~ \distance_\preshapes(\surfaceone, \rotation( \surfacetwo \circ \diffeo )). 
  \label{eq:spatialregisterationsurfacespace}
\end{equation}

\noi The metric $\distance_\preshapes$ should measure the amount of bending and stretching one needs to apply to $\surfacetwo$ in order to align it onto $\surfaceone$. Such nonlinear metric, however, is complex to work with since it leads to  a nonlinear optimization. Jermyn \etal~\cite{jermyn2017elastic}
showed that by mapping the surfaces to the space of Square Root Normal Fields (SRNFs), the elastic metric becomes an $\ltwo$ metric. Formally, the SRNF map $\srnfmap$ of a surface $\surface$ is given by:
\begin{equation}
    \srnfmap(\surface) = \srnf, \text{such that  } \srnf(\pointonsphere) =\frac{\normalfield(\pointonsphere)}{\sqrt{\parallel \normalfield(\pointonsphere) \parallel_2}},
  \label{eq:srnf}
\end{equation}

\noi where $\normalfield$ is the normal field of the surface $\surface$ and is computed directly from the SNS representation using  Eqn.~\eqref{eq:normal_field}.

\begin{figure}[t]
  \centering
   \includegraphics[trim=0 50 0 40, clip, width=1\linewidth]{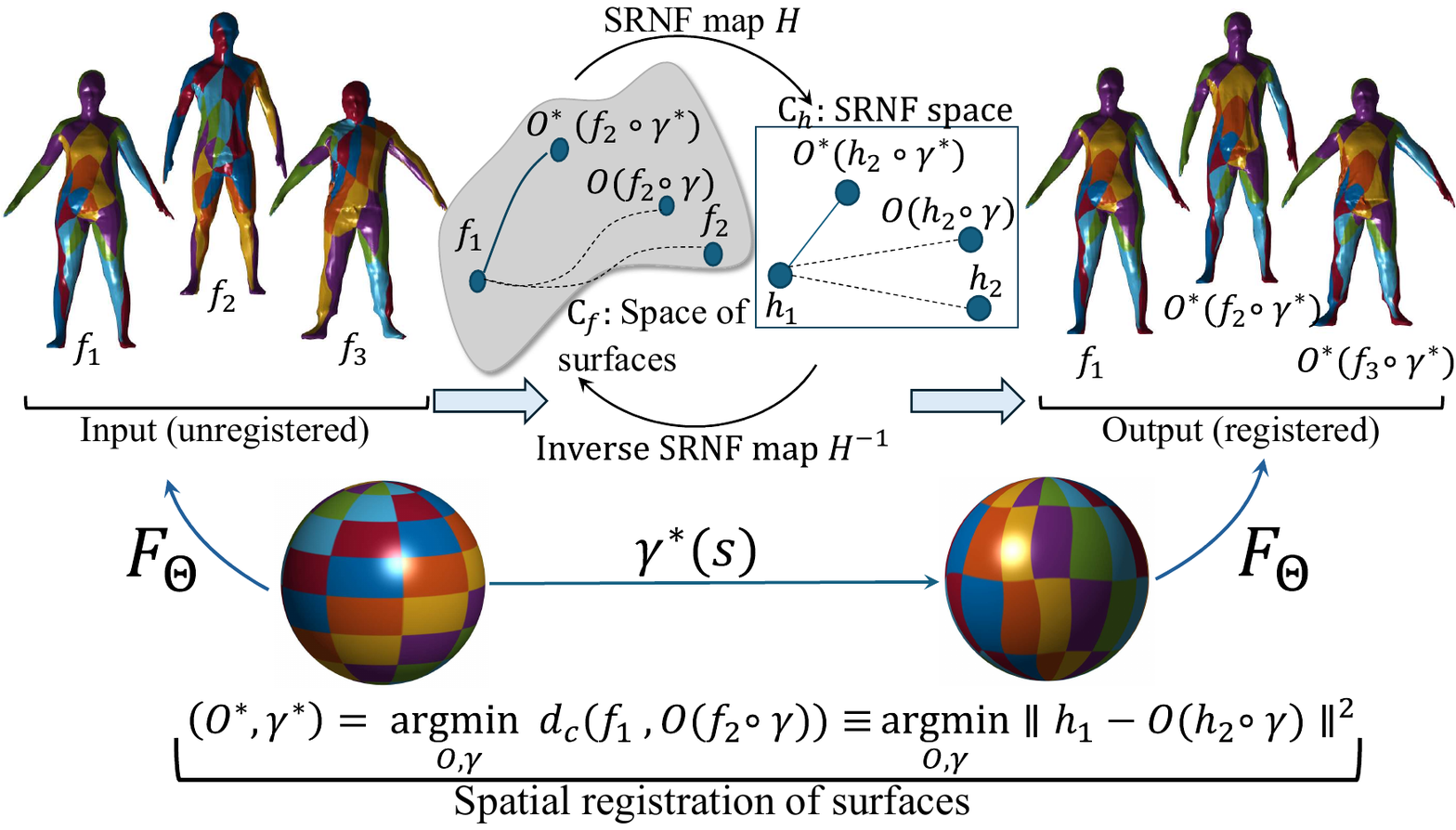}

   \caption{Illustration of the spatial registration of Dynamic Spherical Neural Surfaces (D-SNSs). First,  input D-SNSs are mapped to their SRNF representation. We then use the $\ltwo$ metric in the SRNF space to elastically register the two surfaces, \ie finding the optimal rotation $\rotation^*$ and diffeomorphism $\diffeo^*$ that minimize the $\ltwo$ metric in the SRNF space.}
   \label{fig:spatialregistration}
\end{figure}

\begin{figure}[t]
  \centering
   \includegraphics[trim=240 180 240 180, clip, width=1\linewidth]{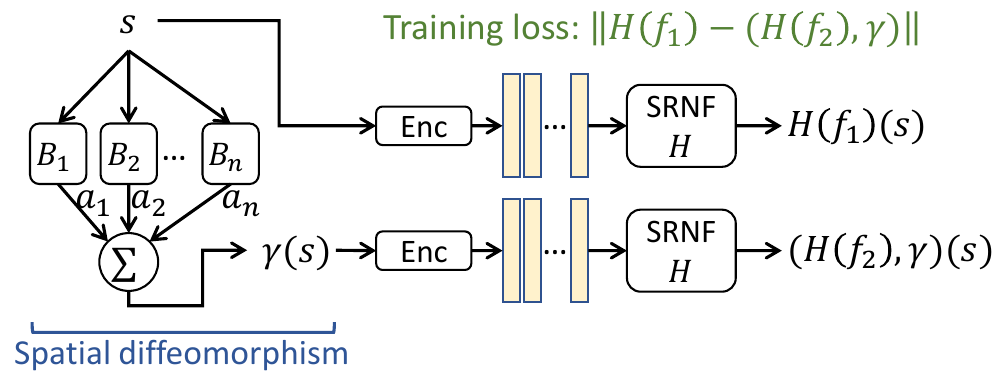}

   \caption{A spatial diffeomorphism can be formulated as an MLP with two layers:  the input layer evaluates the harmonic basis at the query point  $\pointonsphere\in \stwo$. The output layer then computes their weighted sum. The learnable parameters are the weights of the output layer. To perform spatial registration, we freeze the weights of the D-SNS networks and only optimize the weights of the diffeomorphism network using the loss of Eqn.~\eqref{eq:spatialregistrationsrnfspace}.}
   \label{fig:spatial_diffeomorphism}
\end{figure}

With this representation, the spatial registration in Eqn.~\eqref{eq:spatialregisterationsurfacespace} can be reformulated under the SRNF as an optimization problem of the form (see Figure~\ref{fig:spatialregistration}): 
\begin{equation}
  (\rotation^*,\diffeo^*) = \underset{\rotation,\diffeo}{\argmin} \parallel \srnfone - \rotation( \srnftwo \circ \diffeo ) \parallel^2. 
  \label{eq:spatialregistrationsrnfspace}
\end{equation}

\noi Here,  $\srnfone = \srnfmap(\surfaceone)$ and $\srnftwo = \srnfmap(\surfacetwo)$. To solve this optimization problem, we represent the diffeomorphism $\diffeo$, which is a function on $\stwo$, using the weighted sum of spherical harmonic basis: $\diffeo = \weightedharmbasis$ and estimate the optimal weights $\{\harmweights\}$ using gradient descent; see Figure~\ref{fig:spatial_diffeomorphism}.  During training, we freeze the weights and jointly optimize for rotation and diffeomorphism in an iterative manner until convergence: we first fix $\diffeo$ and solve for the rotation $\rotation$ using Singular Value Decomposition. We then fix  $\rotation$ and solve for $\diffeo$ by  optimizing  the loss of Eqn.~\eqref{eq:spatialregistrationsrnfspace} using gradient descent. We repeat this process until convergence.

\subsubsection{Temporal registration} 
\label{sec:temporal_registration}
Given two spatially registered 4D surfaces and both represented using their D-SNS $\spatiotemporalsurface_1$ and $\spatiotemporalsurface_2$, their temporal registration can be formulated as that of finding a time-warping function $\timewarp:[0, 1] \to [0, 1]$ that brings  $\spatiotemporalsurface_2$ as close as possible to $\spatiotemporalsurface_1$. This is an optimization problem of the form:
\begin{equation}
  \timewarp^* = \underset{\timewarp}{\argmin} ~\distance (\spatiotemporalsurface_1, \spatiotemporalsurface_2 \circ \timewarp). 
  \label{eq:temporalregisterationsurfaces}
\end{equation}

\noi where $\distance$ is a measure of closeness. 
We treat time warping as curve stretching and thus, $\distance$ should measure elasticity. 
However, working directly with the 4D surfaces $\spatiotemporalsurface_1$ and $\spatiotemporalsurface_2$ is computationally expensive since they are of infinite dimension (given the continuous nature of the proposed representation). Also, the metric $\distance$ should measure elasticity and thus is nonlinear, which makes the optimization problem of Eqn.~\eqref{eq:temporalregisterationsurfaces} complex to solve.


To address the first problem, we map the 4D surfaces to a low dimensional space of curves using Principle Component Analysis. This way, a 4D surface $\spatiotemporalsurface$ becomes a curve  $\pcacurve$ in the low-dimensional PCA space. Now, to address the second problem, we further map the curve to their Square Root Velocity Fields (SRVFs). Srivastava \etal~\cite{srivastava2010shape} showed that the partial elastic metric that measures bending and stretching of curves  becomes an $\ltwo$ metric in the SRVF space. Formally, the SRVF of a curve $\pcacurve$ is defined as its tangent field scaled by the square root of the tangent field, \ie:
\begin{equation}
    \srvf = \srvfmap(\pcacurve) = \frac{\srvftangentfield }   {\sqrt{ \parallel \srvftangentfield \parallel}}. 
  \label{eq:temporalSRVFs}
\end{equation}

\begin{figure}[t]
  \centering
   \includegraphics[trim=0 65 0 50, clip, width=1\linewidth]{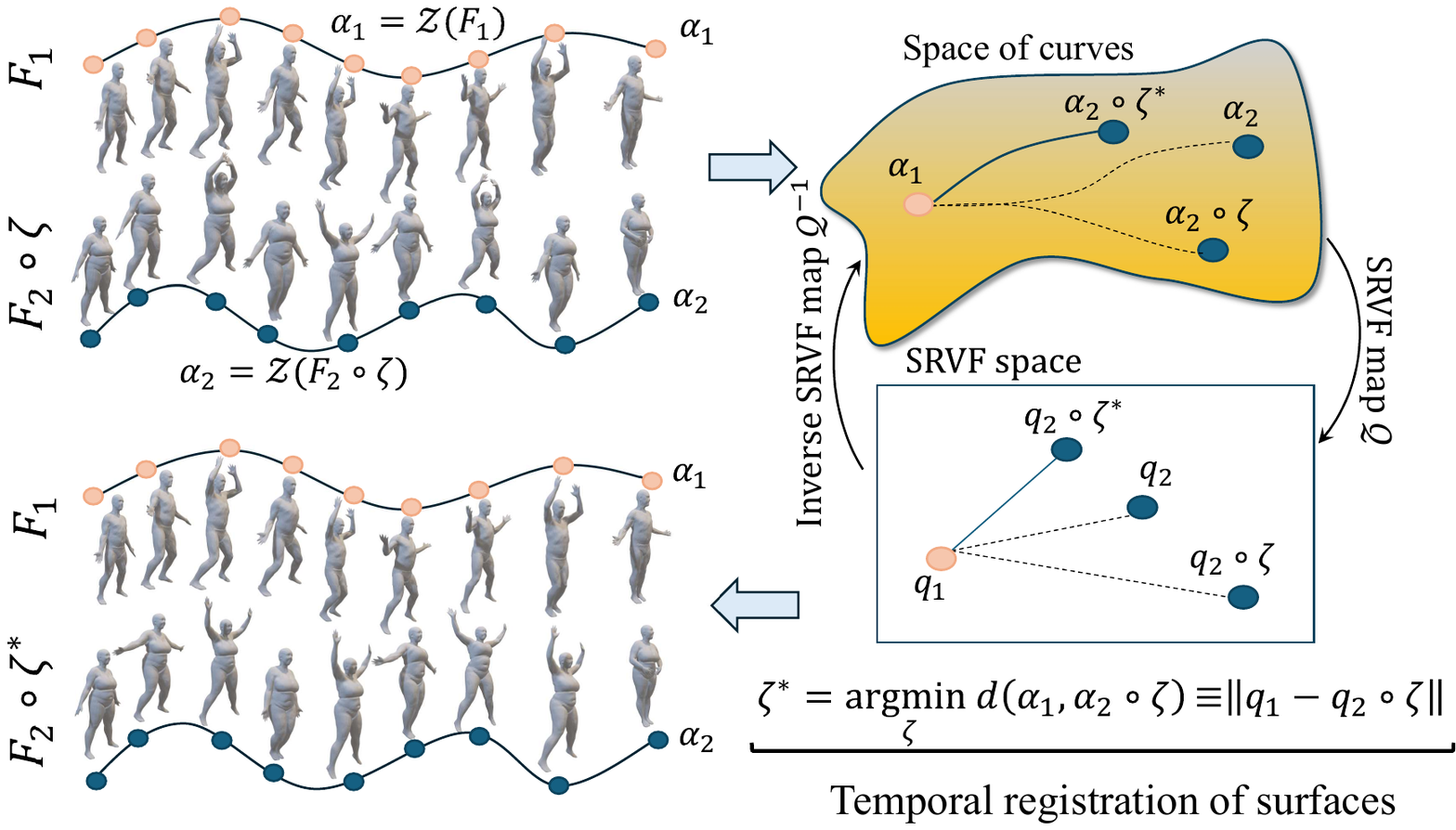}

   \caption{Illustration of the temporal registration process of two 4D surfaces $\spatiotemporalsurface_1$ and $\spatiotemporalsurface_2$ using the proposed Dynamic Spherical Neural Surfaces representation. The 4D surfaces are first mapped to the SRVF space, which has an Eucldiean structure. Thus, we formulate the temporal registration of 4D surfaces as that of elastic registration of curves in the SRVF space.}
   \label{fig:temporalregistration}
\end{figure}

\noi Here,  $\srvftangentfield = \frac{\partial\pcacurve}{\partial\thetime}$ is the tangent vector field on $\pcacurve$. Given two trajectories $\pcacurveone$ and $\pcacurvetwo$, we represent them in their respective SRVFs $\srvfone$ and $\srvftwo$. Then, Eqn.~\eqref{eq:temporalregisterationsurfaces} is reformulated as:
\begin{equation}
  \timewarp^* = \underset{\timewarp}{\argmin} 
  \parallel \srvfone - \srvftwo \circ \timewarp\parallel^2. 
  \label{eq:temporalregisterationsrvfs}
\end{equation}

\noi We treat $\timewarp$ as a continuous function and implement it using an MLP that takes time $\thetime \in \real$ and returns $\timewarp(\thetime)$, which we enforce to in $[0, 1]$ by applying to it  a Sigmoid activation function.   During training, we freeze the parameters of the networks of $\srvfone$ and $\srvftwo$ and only update the parameters of $\timewarp$  by minimizing the loss of Eqn.~\eqref{eq:temporalregisterationsrvfs}. However, to ensure that the MLP learns a diffeomorphism in the temporal domain $[0,1]$, \ie, $ \timewarpnetwork : [0,1] \to [0,1]$ such that  $\timewarpoutputdomain$, we regularize the time-warping network by enforcing the   first derivative of the network with respect to time $t$ to be non-negative. This is done by adding the following regularization term to the loss function:
\begin{equation}
  \monotonicityregularization = \int_0^1 \max\left(0, -\frac{\partial \timewarpnetworkwithoutparameters(t)}{\partial t} \right).
  \label{eq:temporalregisterationmonotonic}
\end{equation}

\noi Hence, the total loss for temporal registration is:
\begin{equation}
  L = \parallel \srvfone - \srvftwo \circ \timewarp\parallel^2  + \lambda \monotonicityregularization.
  \label{eq:temporalregisterationtotal}
\end{equation}

\noi Figure~\ref{fig:temporalregistration} summarizes the temporal registration process. We use an MLP composed of two residual blocks. Each block has two layers of $32$ neurons each. 
We train the network using mini-batches of discretized time intervals and change the input samples after every $200$ epochs to learn a continuous temporal registration.   


\subsection{Geodesics}
\label{sec:geodesics}
Let $\spatiotemporalsurface_1$ and $\spatiotemporalsurface_2$ be two 4D surfaces after their spatiotemporal registration, using the approach described in Section~\ref{sec:registration}. The geodesic between the two 4D surfaces can now be defined as the shortest path, with respect to the elastic metric of Eqn.~\eqref{eq:spatiotemporal_registration_loss}. However, instead of working with this complex metric, we first map the 4D surfaces to the SRVF space, following the approach described in Section~\ref{sec:temporal_registration}. Let $\srvfone $ and $\srvftwo$ be the SRVFs of $\spatiotemporalsurface_1$ and $\spatiotemporalsurface_2$, respectively. Since the SRVF space is Euclidean, then the geodesic path $\geod_\srvf$ between $\srvfone $ and $\srvftwo$ is a straight line, \ie
\begin{equation}
  \geod_\srvf(\variation) = (1 - \variation)\srvfone + \variation\srvftwo, \quad  \variation \in [0,1], 
  \label{eq:geodesics}
\end{equation}

\noi which is straightforward to compute.  To visualize the geodesic paths $\geod_{\srvf}$, we first map the obtained geodesics in the SRVF space back to the space of 4D surfaces through SRVF inversion, which has an analytical form~\cite{srivastava2010shape},  followed by PCA inversion. Figure~\ref{fig:results4dgeodesicshuman} shows an example of a geodesic between two 4D surfaces corresponding to 4D human performing jumping actions.

\subsection{Co-registration and Karcher mean}
\label{sec:statistics}



The mean 4D surface $\meanfourDsurf$, also called Karcher mean, of a set 4D surfaces $\setofsurfaces$ is the 4D surface that is the closest, with respect to the metric of Eqn.~\eqref{eq:spatiotemporal_registration_loss},  to all the input 4D surfaces after their spatiotemporal co-registration. Thus, the joint problem of computing the mean 4D surface and co-registration is  an optimization problem of the form:  
\begin{gather}
    \nonumber \meanfourDsurf, \{\rotation_i^*, \diffeo_i^*, \timewarp_i^*\}_{i=1}^{n} = \\
    \underset{\spatiotemporalsurface, \{\rotation_i, \diffeo_i, \timewarp_i\}_{i=1}^{n}}{\argmin} \sum_{i=1}^n \int_0^1 \loss\left(\spatiotemporalsurface(\thetime),  \rotation(\spatiotemporalsurface_i \circ \timewarp)(\thetime) \circ  \diffeo\right)\mathrm{d}\thetime.
    \label{eq:karcher_mean}
\end{gather}

\noi Similar to the pairwise temporal registration, we solve this optimization in two stages. First, we separately perform the spatial registration of the input 4D surfaces by computing $(\rotation_i^*, \diffeo_i^*)$ following the approach described in Section~\ref{sec:spatial_registration}. Then, we map the spatially registered 4D surfaces to their SRVF space and compute the mean  as well as the  temporal warpings $\{\timewarp_i^*\}_{i=1}^{n}$ using the $\ltwo$ metric, which is equivalent to the elastic metric $\loss$ in the original space, \ie: 
\begin{equation}
  \optimizedsrvf, \{\timewarp_i^*\}_{i=1}^{n} = \underset{\srvf,  \{\timewarp_i\}_{i=1}^{n}}{\arg\min} \underset{i=1}{\overset{n}{\sum}} \parallel \srvf - \srvf_i \circ \timewarp_i \parallel^2.
  \label{eq:meanstatistics}
\end{equation}

\noi Once solved, the mean 4D surface $\meanfourDsurf$ can be  computed analytically by applying the inverse SRVF~\cite{srivastava2010shape} to $\optimizedsrvf$. 




\section{Results}
\label{sec:results}
\textbf{Implementation details.} We trained our neural networks using NVIDIA GeForce RTX $4090$ GPU with $2.4$ GHz Intel Core i9. All networks are optimized using RMSProp with a learning rate of $10^{-4}$ and a momentum of $0.9$. We trained D-SNS for  $50$K epochs, which takes between two to three hours depending on the number of vertices in the mesh, the spatial diffeomorphism for $500$ iterations, which takes around  two minutes, and the time warping network for up to $3,000$ epochs, which takes between $10$ to $30$ minutes. The D-SNS and the time-warping network use SoftPlus activation. We initialize the weights of the time-warping network by overfitting it to the identity diffeomorphism. 



\vspace{3pt}
\noi\textbf{Datasets. } We have evaluated the proposed framework on the 4D face and 4D human body surfaces from the MPI DFAUST~\cite{dfaust:CVPR:2017},  VOCA~\cite{VOCA2019}, COMA~\cite{COMA:ECCV18}, and  CAPE~\cite{CAPE:CVPR:20}; see the Supplementary Mayerial. The datasets come with registered triangulation meshes. We spherically parameterize these datasets using the implementation~\cite{kurtek2013landmark} of  Praun and Hoppe's approach~\cite{praun2003spherical}. We then generate random diffeomorphisms to simulate non-registered surfaces.


\begin{table}[t]
    \centering
    \resizebox{\linewidth}{!}
    {
    \begin{tabular}{@{}lccc@{}}
        \toprule
         \textbf{Datasets} & \textbf{Mean $(\times 10^{-6})$} & \textbf{Std $( \times 10^{-6})$} & \textbf{Median $( \times 10^{-6})$} \\
       \hline
        DFAUST~\cite{dfaust:CVPR:2017} & 1.60 & 0.52 & 1.52  \\ \hline
        CAPE~\cite{CAPE:CVPR:20} & 1.51 & 0.68 & 1.27  \\ \hline
        COMA~\cite{COMA:ECCV18} & 2.29 & 1.68 & 1.66  \\ \hline
        VOCA~\cite{VOCA2019} & 0.89 & 0.51 & 0.79  \\ 
        \bottomrule
    \end{tabular}
    }
    \caption{Evaluation of the performance of the proposed neural representation on four  datasets. We use  the average  $\ltwo$ distance between the ground truth discrete and the neural surface.  }
    \label{tab:neuralrepresentation}
\end{table}

\begin{table}[t]
    \centering
    \resizebox{\linewidth}{!}
    {
    \begin{tabular}{@{}l|ccc|ccc@{}}
        \hline
          & \multicolumn{3}{c|}{\textbf{Original temporal samples}} & \multicolumn{3}{c@{}}{\textbf{$30$ temporal samples}} \\
        \cline{2-7}
         \textbf{Datasets} & \textbf{Mean } & \textbf{Std } & \textbf{Median } & \textbf{Mean } & \textbf{Std } & \textbf{Median }\\
        & $( \times 10^{-5})$ & $( \times 10^{-5})$ &$( \times 10^{-5})$ & $( \times 10^{-5})$ & $( \times 10^{-5})$ &$( \times 10^{-5})$ \\ 
       \hline
        DFAUST~\cite{dfaust:CVPR:2017} & 0.17 & 0.08 & 0.16 & 1.00 & 0.55 & 1.20\\ \hline
        CAPE~\cite{CAPE:CVPR:20}  & 0.14 & 0.02 & 0.12 & 1.34 & 0.72 & 1.25\\ \hline
        COMA~\cite{COMA:ECCV18} & 0.23 & 0.10 & 0.20 & 0.14 & 0.02 & 0.14 \\ \hline
        VOCA~\cite{VOCA2019}  & 0.10 & 0.04 & 0.10 & 0.11 & 0.06 & 0.08\\ 
        \bottomrule
    \end{tabular}
    }
    \caption{D-SNS interpolation results on original and $30$ temporal samples, computed using the average $\ltwo$ distance between the ground truth and the neural surfaces. Results are shown for five random 4D surfaces across four datasets. Note that, due to small errors, all values are shown as multipliers of  $10^{-5}$. }
    \label{tab:neuralrepresentationinterp}
\end{table}

\subsection{Dynamic Spherical Neural Surfaces results}
\label{dsns-results}
 In this experiment, we take the ground truth discrete 4D surfaces of DFAUST, COMA, CAPE, and VOCA datasets and train one D-SNS per 4D surface. To quantitatively evaluate the quality of the neural representation, we measure the average point-wise $\ltwo$ distance between the ground truth discrete 4D surface and the D-SNS representation. Table~\ref{tab:neuralrepresentation} summarizes the mean, median, and standard deviation of the representation error over each of the four datasets. As one can see, the errors are in the order of  $10^{-6}$, which demonstrates that the novel neural representation can faithfully represent genus-0 4D surfaces. (Note that all the surfaces are normalized for translation and scale and thus they fit within a bounding sphere of radius $1$.) 

\begin{figure}[t]
  \centering

    \includegraphics[trim=0 15 20 0, width=.98\linewidth]{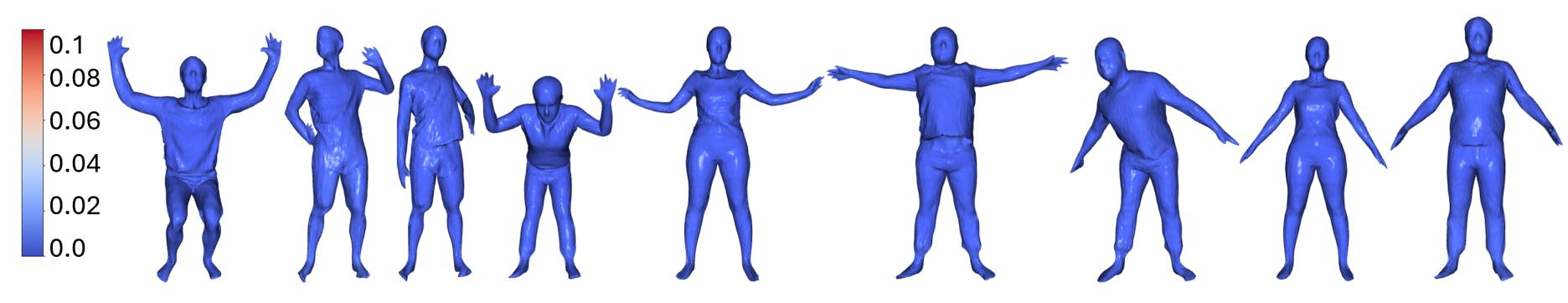}\\
    (a) 4D humans.
    \includegraphics[trim=0 15 20 0, width=\linewidth]{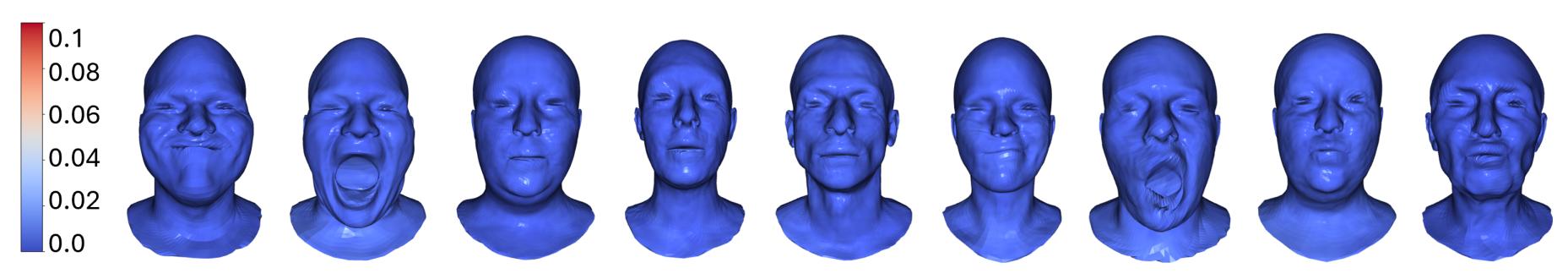}\\
    (b) 4D faces.


   \caption{We measure the pointwise error between the proposed D-SNS and the groundtruth discrete 4D surfaces. Here, we show some time frames with the error plotted as a heatmap. Observe that the proposed neural representation can accurately represent 4D surfaces, with a pointwise error that is less than $0.01$. Please refer to the Supplementary Material for additional results.}
   \label{fig:resultsneuralrepresentation1}
\end{figure}

Figure~\ref{fig:resultsneuralrepresentation1} shows some results demonstrating the representation quality of the proposed D-SNS. In this figure, we picked up some surfaces and showed, in the form of a heat map,  the error between the original surface and the neural surface. As one can see, the errors are less than $0.01$. This shows that the proposed D-SNS can accurately represent complex 4D surfaces. Table~\ref{tab:neuralrepresentationinterp}, on the other hand, demonstrates its interpolation ability. In this experiment, the neural representation was trained only on $30$ temporal samples of the entire sequence, yet, the method is able to interpolate those keyframes and generate plausibly  smooth 4D surfaces. This demonstrates the D-SNS ability to successfully interpolate temporally-sparse sequences.

\subsection{Spatiotemporal registration and geodesics}

\begin{figure}[t]
  \centering
   \includegraphics[width=1\linewidth]{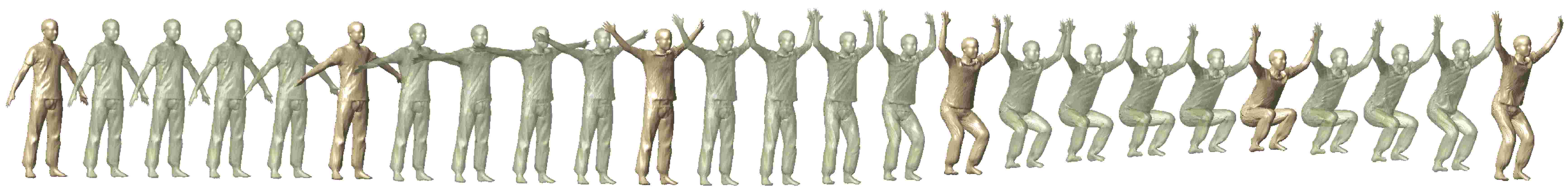}\\
   (a) Source.\\
   \includegraphics[width=1\linewidth]{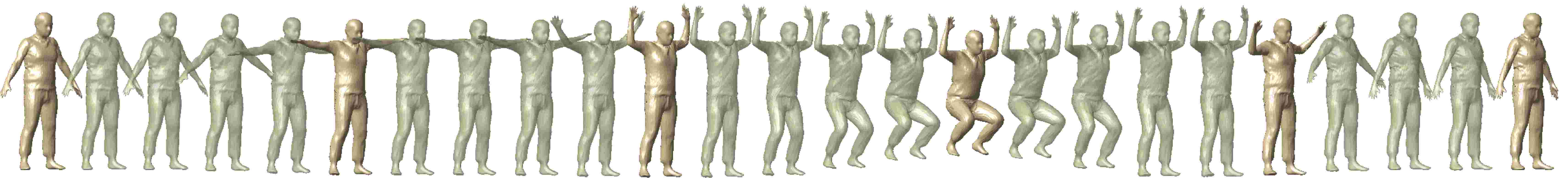}\\
   (b) Target before spatiotemporal registration.\\
    \includegraphics[width=1\linewidth]{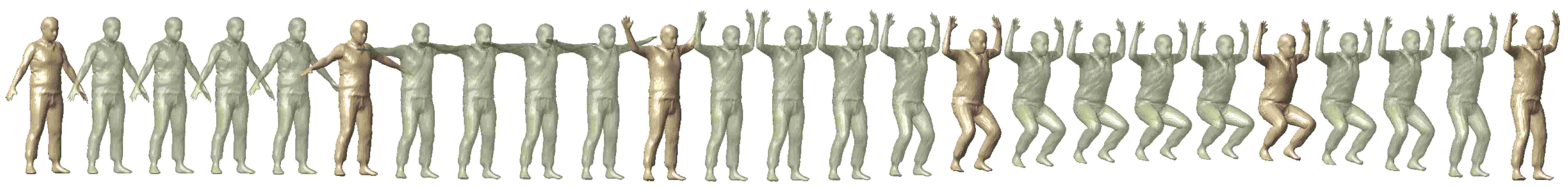}\\
   (c) Target after spatiotemporal registration.\\
   
   \caption{Example of the spatiotemporal registration of two 4D humans from the CAPE dataset. The proposed framework aligns the target 4D surface \textbf{(b)}  onto the source \textbf{(a)}. The target neural surface after registration \textbf{(c)} is fully aligned with the source. Please refer to the Supplementary Material for additional results.}
   \label{fig:resultsspatiotemporalbody}
\end{figure}

Figure~\ref{fig:resultsspatiotemporalbody} shows an example of the spatiotemporal registration of pairs of 4D  humans deforming at different rates. We show the 4D surfaces before and after their temporal registration. As we can see, our approach is capable of bringing the sequences in full alignment. Note that the spatiotemporal registration of human bodies is more complex than faces due to greater motion and body articulation, which leads to significant bending and stretching. Please refer to the Supplementary Material for additional visual results.

We quantitatively evaluate the proposed temporal registration against the method of Laga \etal~\cite{laga20234datlas} using the  evaluation framework proposed in 4D Atlas~\cite{laga20234datlas}, since, to the best of our knowkedge, is the only method paper that dealt with the analysis of 4D surfaces. We take a   4D neural surface $\spatiotemporalsurface_i$ and apply to it  $10$ random temporal diffeomorphisms $\timewarpnetworkwithoutparameters^*_{ij}$ resulting in $10$ surfaces $\{\spatiotemporalsurface_j\}_{j=1}^{10}$  that differ in their execution speeds. Then, for each pair $(\spatiotemporalsurface_i, \spatiotemporalsurface_j)$, we compute, using the proposed approach, the optimal time warping  $\timewarpnetworkwithoutparameters^*_{ij}$ that aligns $\spatiotemporalsurface_j$ onto $\spatiotemporalsurface_i$. Ideally, $\timewarpnetworkwithoutparameters^*_{ij}$  should be the inverse of $\timewarpnetworkwithoutparameters_{ij}$. Thus, we use the distance between $\timewarpnetworkwithoutparameters^*_{ij}$ and the inverse of $\timewarpnetworkwithoutparameters_{ij}$ as a measure of error, which we aggregate over the $10$-random diffeomorphisms. 

\begin{figure}[t]
  \centering  
  
   \includegraphics[width=.90\linewidth]{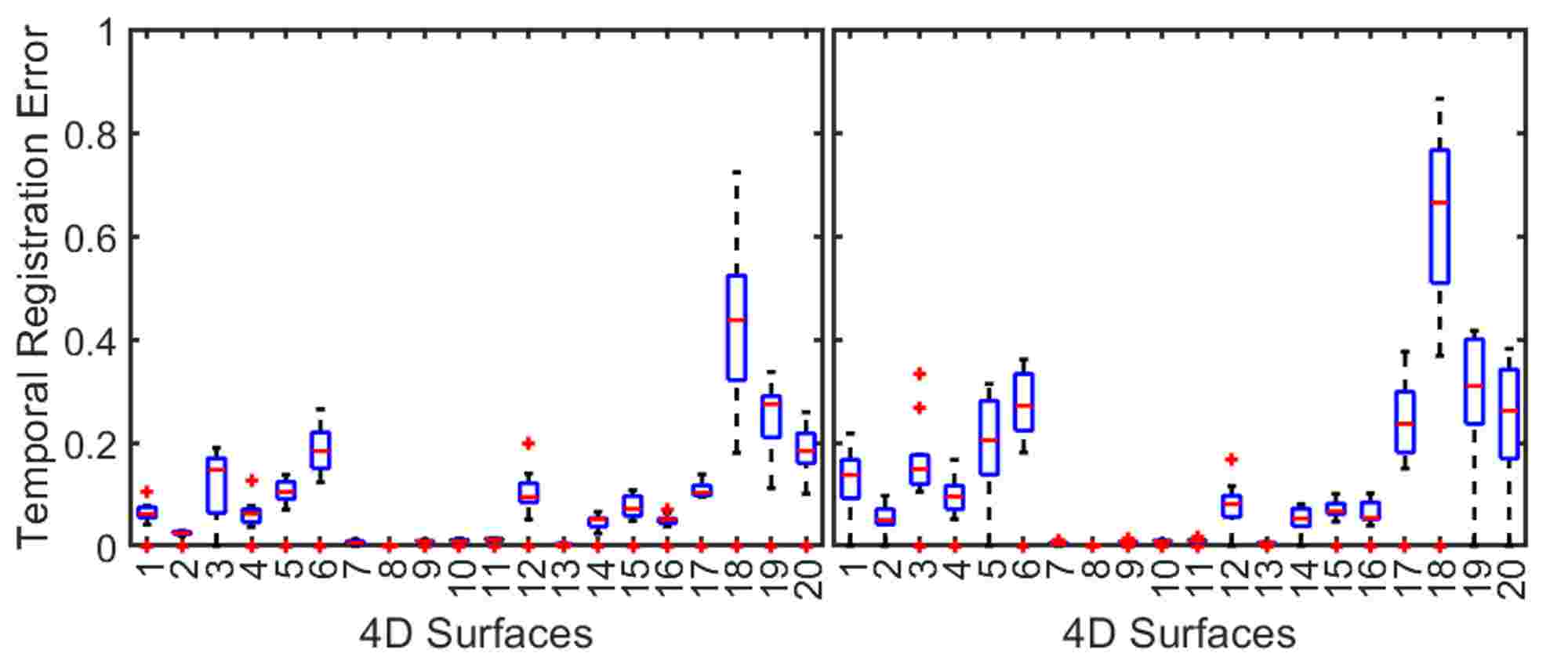}\\
    \begin{minipage}{0.60\columnwidth}
        \centering
        50 samples
    \end{minipage}%
    \begin{minipage}{0.325\columnwidth}
        \centering
        25 samples
    \end{minipage} \\
    \textbf{(a) 4D Atlas~\cite{laga20234datlas}.} \\

   \includegraphics[width=.90\linewidth]{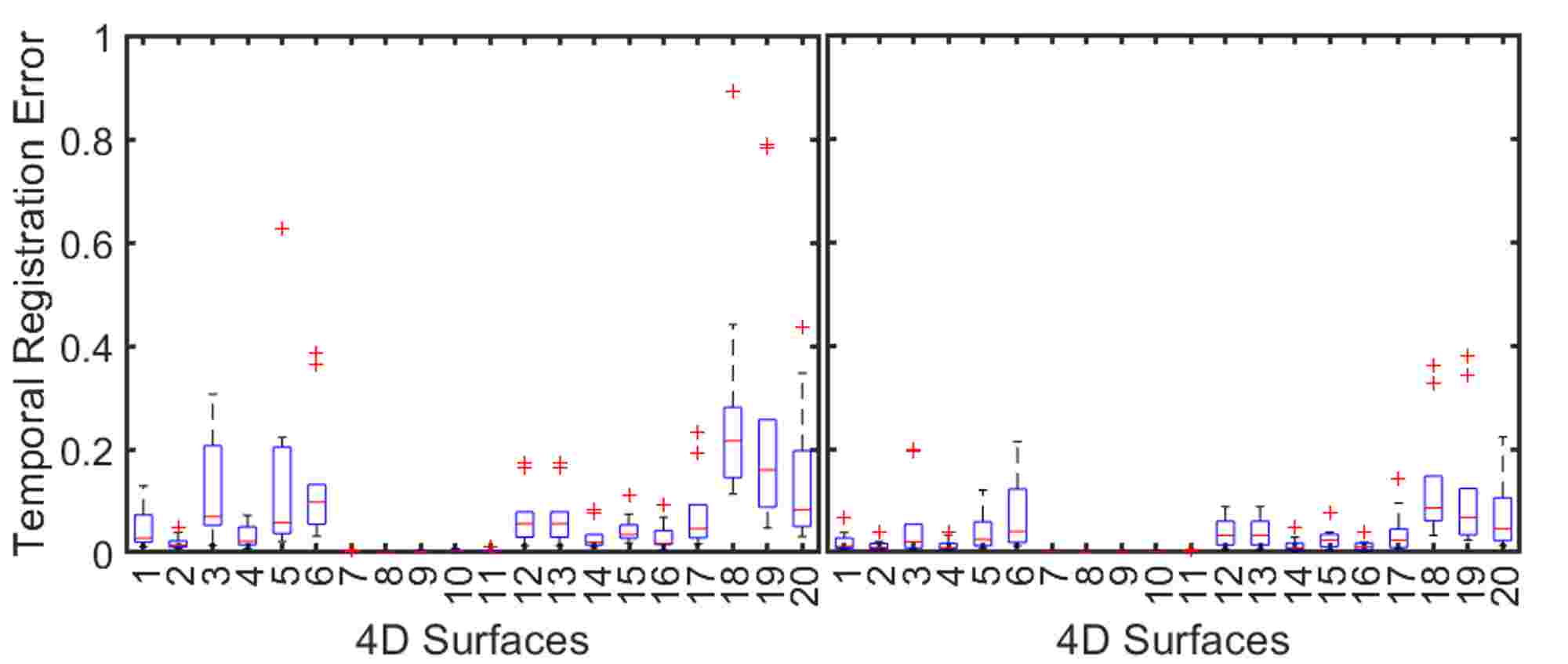}
    \begin{minipage}{0.60\columnwidth}
        \centering
        50 samples
    \end{minipage}%
    \begin{minipage}{0.325\columnwidth}
        \centering
        25 samples
    \end{minipage} \\
    \textbf{(b) Ours.}    
   \caption{Boxplots providing the statistics on the accuracy of the temporal registration when using  two different temporal samplings ($50$ and $25$ temporal samples). We run the experiment on $20$ pairs, each pair is randomly sampled with $10$ different random temporal diffeomorphisms and report the statistics on the temporal registration error aggregated over the $10$ runs.  The red lines denote the median error, while the boxes represent its spread.}
   \label{fig:evaluationboxplot}
\end{figure}

We perform this experiment on the four datasets,   using our method and 4D Atlas~\cite{laga20234datlas}. We use two different temporal samplings  ($50$ and $25$ temporal samples) to analyze the robustness of both methods. Figure~\ref{fig:evaluationboxplot} presents the results in the form of box plots. This experiment shows that when the temporal sampling is dense, the two methods achieve similar performance, with a slight advantage to ours. However, when we reduce the temporal sampling ($25$ samples), we observe that the alignment error of 4D Atlas~\cite{laga20234datlas} increases while ours remains more or less stable. This is due to the fact that we treat 4D surfaces as continuous functions  and thus they are less sensitive to the sampling density. 

\begin{table*}[t]
    \centering
    \resizebox{\textwidth}{!}
    {
        \begin{tabular}{@{}l|ccc|ccc|ccc|ccc@{}}
            \hline
                    & & \textbf{CAPE~\cite{CAPE:CVPR:20}} &        & &\textbf{DFAUST~\cite{dfaust:CVPR:2017}} &        & & \textbf{COMA~\cite{COMA:ECCV18}} &        & & \textbf{VOCA~\cite{VOCA2019}} & \\
             \hline
                    & Mean   & Std    & Median & Mean    & Std    & Median & Mean   & Std    & Median & Mean   & Std    & Median \\ 
            \hline
            4D Atlas& 1.2938 & 0.4764 & 1.404  & 2.3785 & 1.1237 & 1.8665 & 0.0597 & 0.0165 & 0.0617 & 0.5026 & 0.1057 & 0.4060  \\
            Ours   & \textbf{0.3574} & 0.1697 & \textbf{0.3199} & \textbf{0.7726} & 0.2037 & \textbf{0.7210} & \textbf{0.0276} & 0.0191 & \textbf{0.0173} & \textbf{0.0994} & 0.0253 & \textbf{0.1030}  \\
    
            \hline
        \end{tabular}
    }
    \caption{Comparison of the proposed spatiotemporal registration with  4D Atlas~\cite{laga20234datlas}. The evaluation is performed on  five pairs performing the same actions. We measure the geodesic distance after the spatiotemporal registration using 4D Atlas~\cite{laga20234datlas} and our approach. The smaller this distance is, the better the alignment. The best results are shown  in \textbf{bold}; see the Supplementary Material for additional results.}
    \label{tab:temporalevaluationmain}
\end{table*}

We also evaluate the registration performance on different 4D surfaces performing the same action; see Table~\ref{tab:temporalevaluationmain}. In this experiement, we take $5$ pairs from each data set, perform their temporal registration using 4D Atlas~\cite{laga20234datlas} and our method. We then measure the geodesic distance between the registered 4D surfaces. As each pair of 4D surfaces perform the same action, the smaller the geodesic distance, the better the alignment. As we can see, our continuous framework is able to outperform 4D Atlas~\cite{laga20234datlas} on all datasets. We performed this experiment using five pairs of 4D surfaces, each pair performing the same action at varying speeds. The Supplementary Material provides more evaluation details.

\begin{figure}[t]
  \centering

    \includegraphics[width=0.97\linewidth]{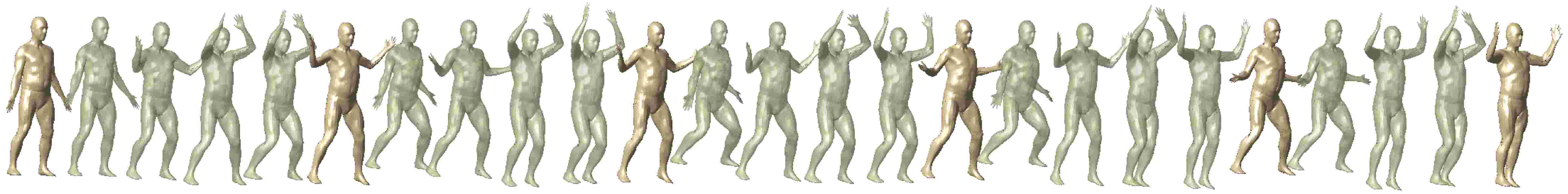}\\
    \includegraphics[width=0.97\linewidth]{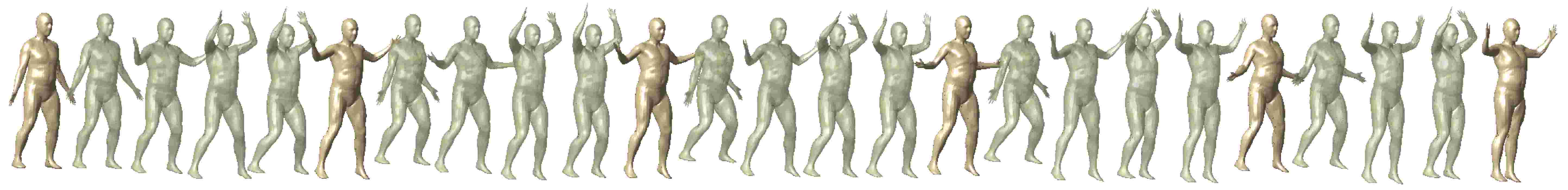}\\
    \fcolorbox{red}{white}{\includegraphics[width=0.97\linewidth]{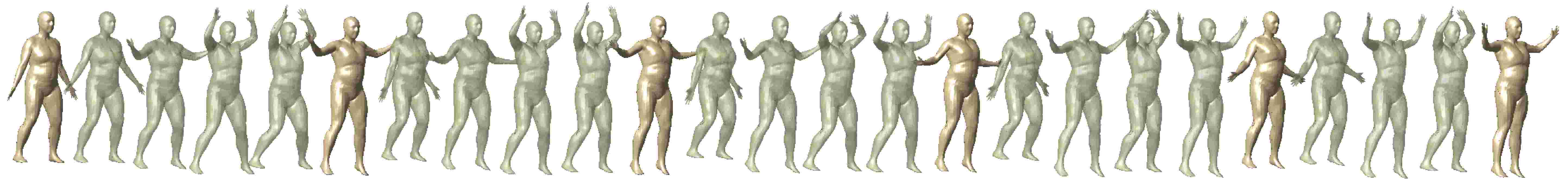}}\\
    \includegraphics[width=0.97\linewidth]{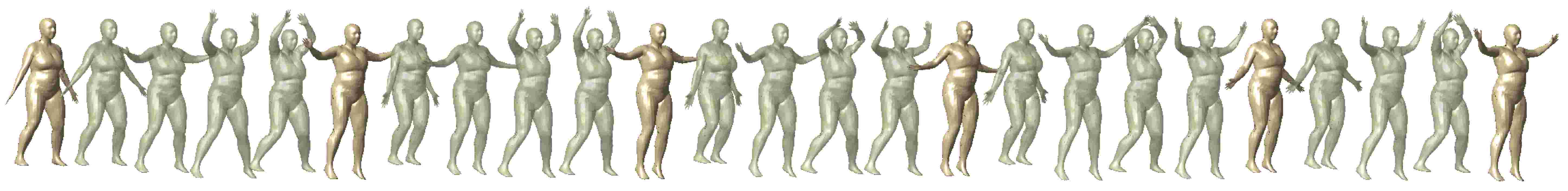}\\
    \includegraphics[width=0.97\linewidth]{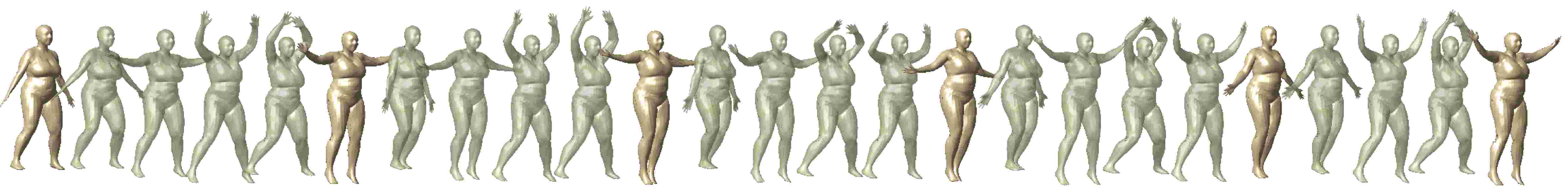}\\

   \caption{ Example of a geodesic after the spatiotemporal registration between two 4D surfaces performing a jumping action.   The first row shows the source 4D surface, the last row shows the target 4D surface, and the intermediate rows show  intermediate 4D surfaces sampled at equidistance along the geodesic path between the source and target. The highlighted row corresponds to the mean 4D surface. The Supplementary Material provides the original 4D surfaces and geodesic  before spatiotemporal registration. The Supplementary Material provides more results.}
   \label{fig:results4dgeodesicshuman}
\end{figure}

\vspace{3pt}
\noi\textbf{4D geodesics.} Figure~\ref{fig:results4dgeodesicshuman} shows an example of a geodesic between 4D neural surfaces performing a jumping action. We show the geodesic  after  spatiotemporal registration of the source (top row) to the target surface (bottom row). The Supplementary Material shows the geodesic before spatiotemporal registration. Observe that before registration (see the Supplementary Material), the jumping actions are misaligned, and the geodesics are not as closely aligned compared to those after registration. We refer the reader to the Supplementary Material for more examples and results.

\begin{figure}[b]
  \centering

    \includegraphics[width=\linewidth]{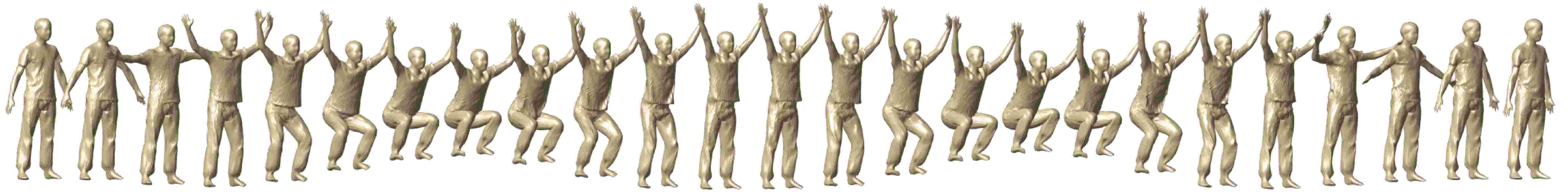}\\
    \includegraphics[width=\linewidth]{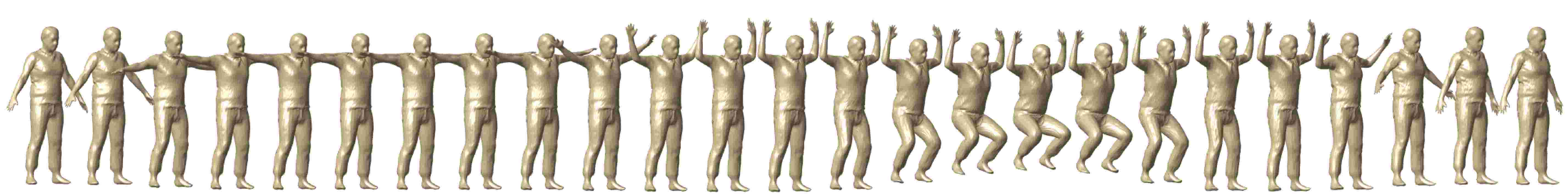}\\
    \includegraphics[width=\linewidth]{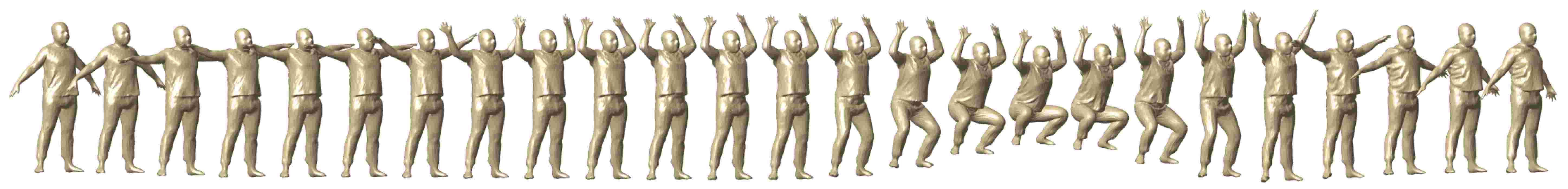}\\
    \includegraphics[width=\linewidth]{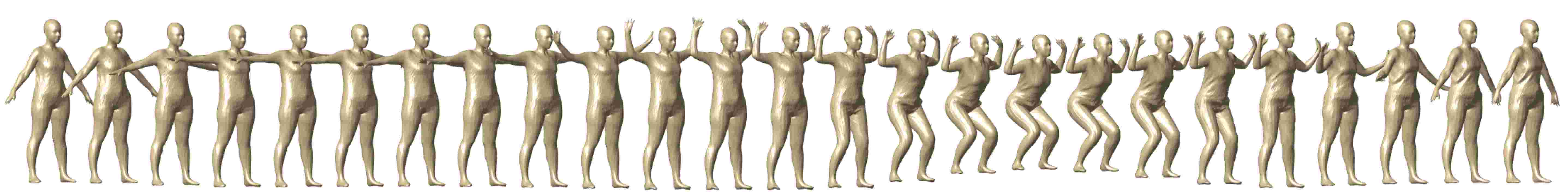}\\
    \includegraphics[width=\linewidth]{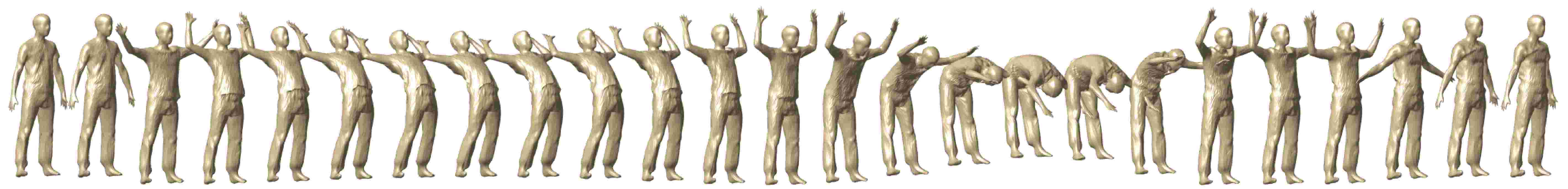}\\
    \includegraphics[width=\linewidth]{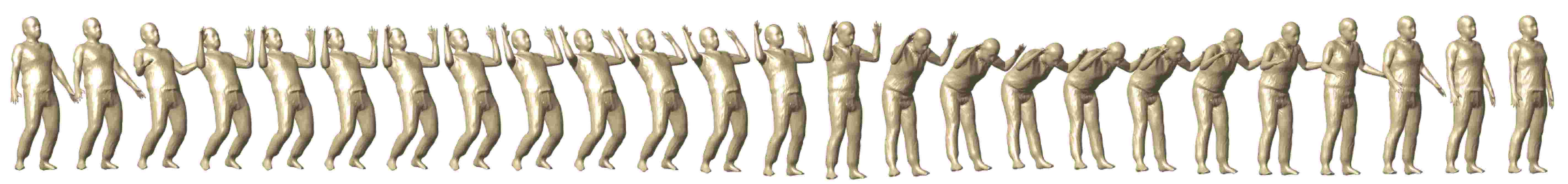}\\
    \fcolorbox{red}{white}{\includegraphics[width=\linewidth]{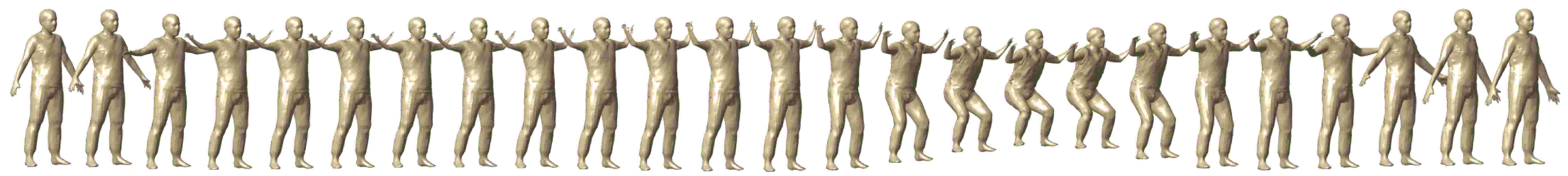}}\\

   \caption{Example of a mean 4D surface (highlighted in red) computed on six 4D surfaces from the CAPE dataset after their spatiotemporal registration. The input 4D surfaces perform different actions:  the 4D surfaces in the first four rows perform a squat action while the last two perform a back and forward bending action; see the Supplementary Material for additional results. 
   }
   \label{fig:results4dmeansurfaces}
\end{figure}

\subsection{Co-registration and mean 4D surfaces}
Figure~\ref{fig:results4dmeansurfaces} shows the 4D mean surface  of a set of  4D neural surfaces, from the CAPE dataset, after their co-registration. These surfaces perform various actions at different execution speeds. The 4D surfaces in the first four rows perform a squat action, and the following two rows perform a bending action. Note that the squat action is repetitive and the number of cycles differs  from one 4D surface to another.  Despite this complexity, the proposed approach is able to co-register the 4D surfaces and compute a plausible 4D mean that is as close as possible to all the other 4D surfaces; see  the Supplementary Material for additional examples.

\vspace{-6pt}
\section{Conclusion}
\label{sec:conclusion}\vspace{-6pt}
We proposed Dynamic Spherical Neural Surfaces as a novel continuous representation of genus-0 4D surfaces that is continuous in both space and time. We then formulated the 4D shape analysis problem as one of analyzing functions. This allows all analysis tasks to be performed directly on these continuous functions without unnecessary discretization. By linking neural networks with the classical Riemannian geometry for 4D shape analysis, this work establishes the first foundational approach that enables comprehensive functional analysis of 4D shapes.

\newpage
{
    \small
    \bibliographystyle{ieeenat_fullname}
    \bibliography{main}
}


\input{supplementary_material}

\end{document}

%% file: notation.tex
\newcommand{\thetime}{t}

\newcommand{\point}{p}

\newcommand{\s}{\ensuremath{\mathbb{S}}}

\newcommand{\spherecoordtheta}{u}
\newcommand{\spherecoordphi}{v}
\newcommand{\stwo}{\ensuremath{\mathbb{S}^2}}

\newcommand{\sn}[1]{\s^{n}}
\newcommand{\domain}{\Omega}					
\newcommand{\pointonsphere}{s}
\newcommand{\singlepointonsphere}{s_i}			

\newcommand{\spatiotemporalsurface}{F} 						
\newcommand{\surface}{f} 	

\newcommand{\meanfourDsurf}{\bar{\spatiotemporalsurface}}

\newcommand{\srnf}{h}							
\newcommand{\srnfone}{\srnf_1}							
\newcommand{\srnftwo}{\srnf_2}							
\newcommand{\srnfmap}{H}                        			

\newcommand{\pointonsurface}{\surfacepoint} 
\newcommand{\surfaceone}{\surface_1} 						
\newcommand{\surfacetwo}{\surface_2} 						

\newcommand{\surfacepoint}{p}
\newcommand{\singlepointonsurface}{\surfacepoint_\surfaceindex}
\newcommand{\predictedsinglepointonsurface}{\surfacepoint_\surfaceindex^*}
\newcommand{\surfaceindex}{i}

\newcommand{\setofsurfaces}{ \{ \spatiotemporalsurface_1, \dots, \spatiotemporalsurface_n \}}

\newcommand{\timewarp}{\timewarpnetworkwithoutparameters}
\newcommand{\timewarpnetwork}{\timewarpnetworkwithoutparameters_\timewarpparameters}

\newcommand{\timewarpnetworkwithoutparameters}{\zeta}
\newcommand{\timewarpparameters}{v}

\newcommand{\timewarpoutputdomain}{0 < \timewarpnetworkwithoutparameters(t) < 1}

\newcommand{\harmweights}{a_i}
\newcommand{\harmbasis}{B_i}
\newcommand{\weightedharmbasis}{\sum_{i=1}^n{\harmweights\harmbasis}}

\newcommand{\neuralrepresentations}{F_\parameters} 				
\newcommand{\neuralrepresentationswithoutparameters}{F}

\newcommand{\parameters}{\Theta}

\newcommand{\pcamap}{\mathcal{Z}}

\newcommand{\variation}{\tau}

\newcommand{\loss}{\mathcal{L}}

\newcommand{\neuralloss}{L_\text{N}}
\newcommand{\monotonicityregularization}{L_\text{M}}

\newcommand{\distance}{d}

\newcommand{\srvf}{q}							
\newcommand{\srvfmap}{Q}                        			
\newcommand{\srvftangentfield}{\overset{.}{\pcacurve}}

\newcommand{\optimizedsrvf}{\overset{-}{\srvf}}

\newcommand{\pcacurveone}{\pcacurve_1}
\newcommand{\pcacurvetwo}{\pcacurve_2}

\newcommand{\srvfone}{\srvf_1}
\newcommand{\srvftwo}{\srvf_2}

\newcommand{\preshapes}{\mathcal{C}}            		
\newcommand{\preshapesf}{\preshapes_{\surface}}        	

\newcommand{\pcacurve}{\alpha}

\newcommand{\normalfield}{\textbf{n}}                   		 

\newcommand{\rotations}{SO(3)}                  			
\newcommand{\rotation}{O}                       				

\newcommand{\diffeos}{\Gamma}                   			
\newcommand{\diffeo}{\gamma}                    			

\newcommand{\geod}{\Lambda}						%

\newcommand{\noi}{\noindent}

\renewcommand{\eg}{\emph{e.g.,\ }}
\renewcommand{\ie}{\emph{i.e.,\ }}

\renewcommand{\etal}{\emph{et al.}}

\usepackage{framed, color}
\definecolor{shadecolor}{rgb}{1,0.5,0}

\newcommand{\real}{\mathbb{R}}

\newcommand{\rcubed}{\real^{3}}
\newcommand{\rthree}{\rcubed}

\newcommand{\ltwo}{\mathbb{L}^{2}}

\def\argmin{\mathop{\rm argmin}}



%% file: supplementary_material.tex
\clearpage
\setcounter{page}{1}


\twocolumn[{%
\renewcommand\twocolumn[1][]{#1}%
\maketitlesupplementary
\centering
\includegraphics[trim=0 180 60 50, width=1\linewidth]{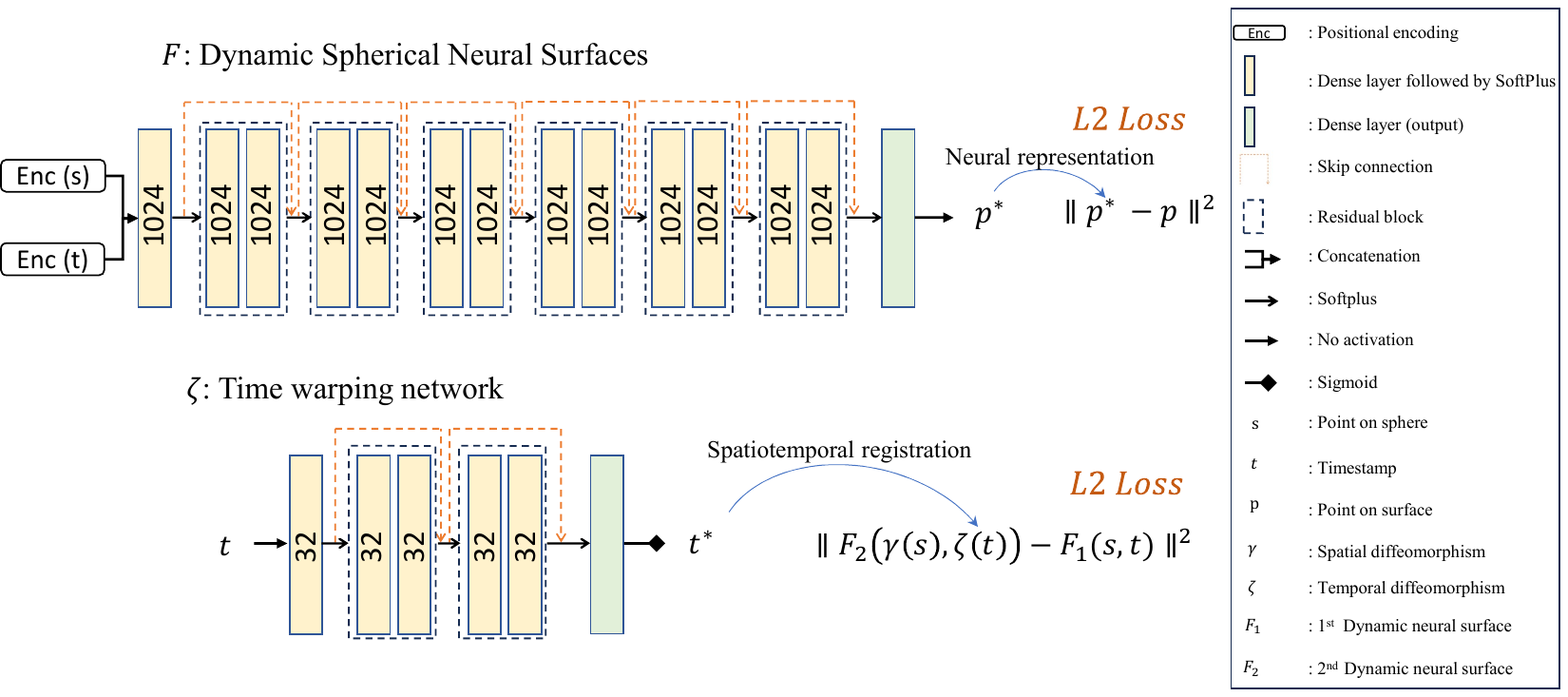}
\captionof{figure}{Detailed architecture of the neural networks used for the Dynamic Spherical Neural Surfaces (D-SNS) and for the time warping.
\vspace{2em}}
\label{fig:network-design}
}]





The Supplementary Material is organized as follows; Section~\ref{suppl:limit} discusses the limitations and potential directions for future work. Section~\ref{suppl:implem} describes the implementation details of our neural framework. Section~\ref{suppl:dataset} includes details on the dataset. Section~\ref{suppl:results} presents additional results of our neural framework. 

\appendix

\begin{figure*}[t]
  \centering

   \includegraphics[trim=0 0 16 0,width=.95\linewidth]{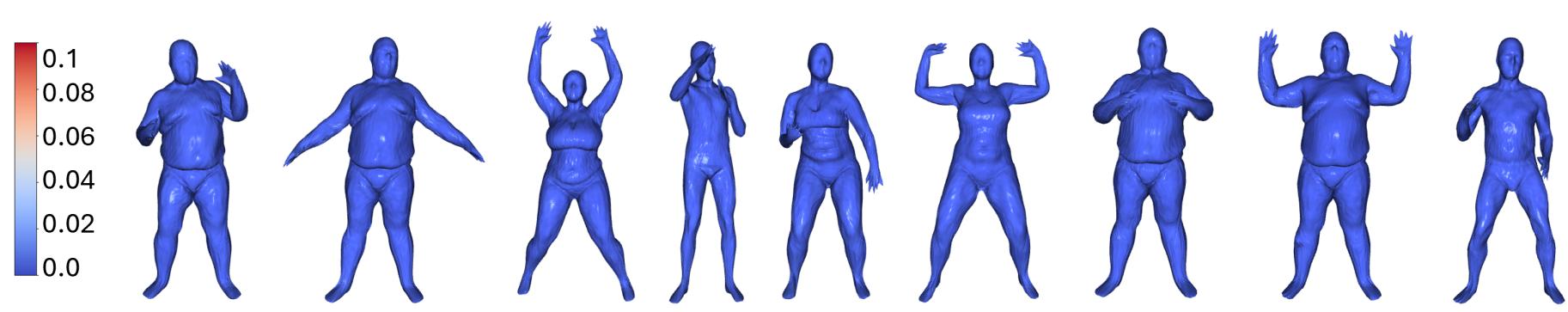}\\
    (a) DFAUST~\cite{dfaust:CVPR:2017}.
    \includegraphics[trim=0 0 20 0, width=.97\linewidth]{figures-arxiv/NeuralRepresentation/CAPE_colbar3.jpg}\\
    (b) CAPE~\cite{CAPE:CVPR:20}.
    \includegraphics[trim=0 0 20 0, width=.97\linewidth]{figures-arxiv/NeuralRepresentation/COMA_colbar3.jpg}\\
    (c) COMA~\cite{COMA:ECCV18}.
    \includegraphics[trim=0 0 20 0, width=.97\linewidth]{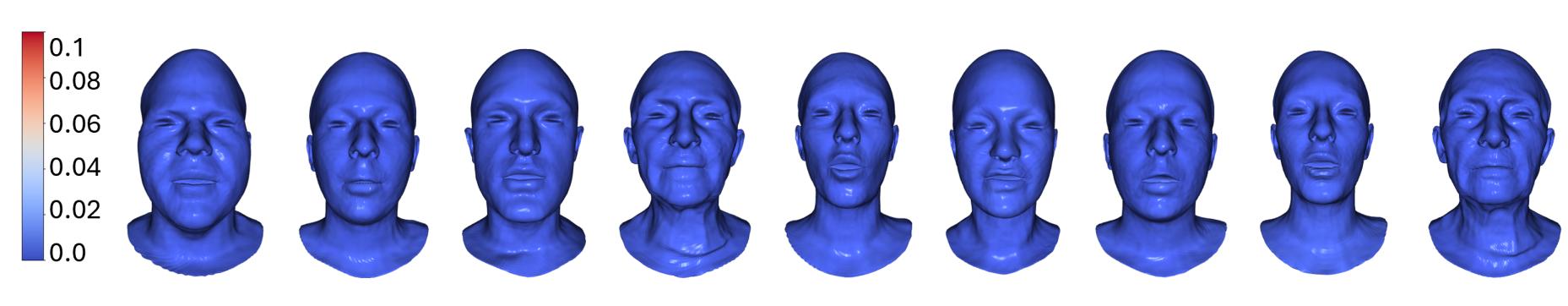}\\
    (d) VOCA~\cite{VOCA2019}.

   \caption{We measure the pointwise error between the proposed D-SNS and the ground truth discrete 4D surfaces. Here, we show some time frames with the error plotted as a heatmap. Observe that the proposed neural representation can accurately represent 4D surfaces, with a pointwise error that is less than $0.01\%$.}
   \label{fig:resultsneuralrepresentation1-suppl}
\end{figure*}

\section{Limitations and future work}
\label{suppl:limit}

\vspace{6pt}
\noi\textbf{Higher genus surfaces.} D-SNS is based on spherical parameterization. Thus, it is limited to  closed genus-0 surfaces. The representation, however, can be easily extended to open genus-0 surfaces, which can be parametrized on open domains such as a disk. Extending the representation to higher genus surfaces would require a complex parameterization, \eg using charts or even a volumetric domain. This will be investigated in future work. 


\vspace{6pt}
\noi\textbf{Computation efficiency.} A key limitation of our neural framework is its high computation time, particularly when training individual D-SNS. Although the representation provides a continuous representation of surfaces, and thus all the  differential properties can be computed analytically, the D-SNS needs to be fitted to every single 4D surface. Thus, it is computationally expensive when analyzing a large number of 4D surfaces. We plan in the future to explore shape-agnostic representations, \eg by following an approach similar to DeepSDF.



\begin{figure*}[t]
  \centering
   \begin{tabular}{c}
    \includegraphics[trim=0 0 25 0,width=1\linewidth]{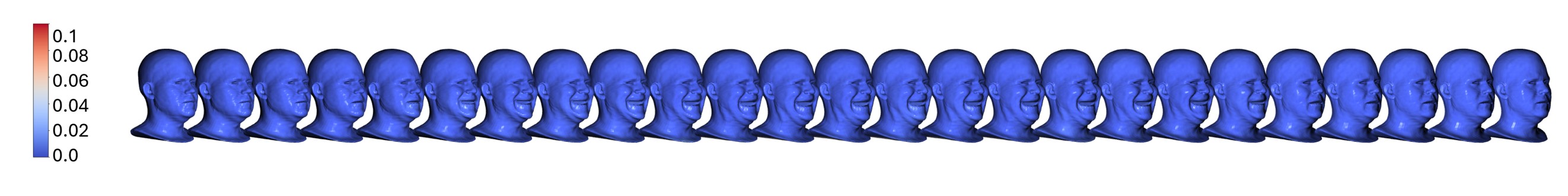}\\
    (a) COMA~\cite{COMA:ECCV18}.\\
    
    \includegraphics[trim=0 0 25 0,width=1\linewidth]{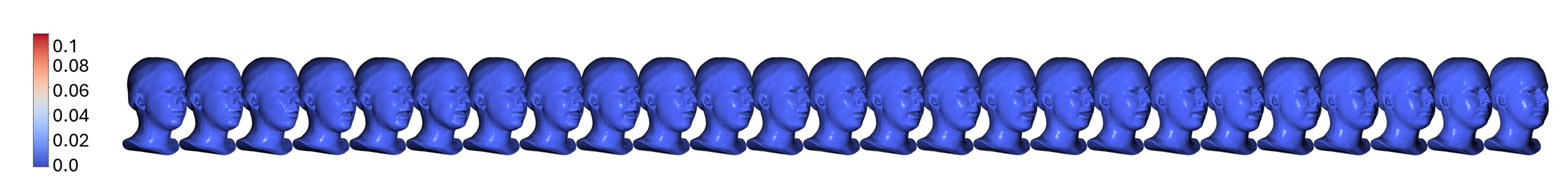}\\
    (b) VOCA~\cite{VOCA2019}. \\
    
    \includegraphics[trim=0 0 25 0,width=1\linewidth]{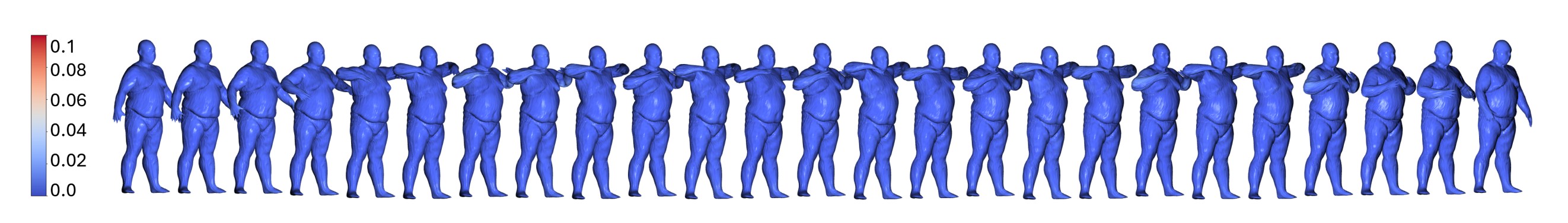}\\
    (c) DFAUST~\cite{dfaust:CVPR:2017}. \\
    
    \includegraphics[trim=0 0 25 0,width=1\linewidth]{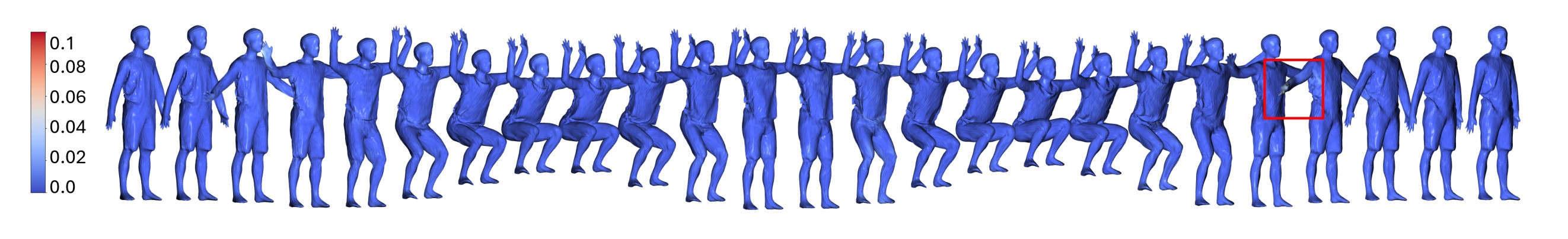}\\
    (d) CAPE~\cite{CAPE:CVPR:20}.

    \end{tabular}

    \begin{tabular}{@{}c@{ }c@{}}    
    \includegraphics[width=0.45\linewidth]{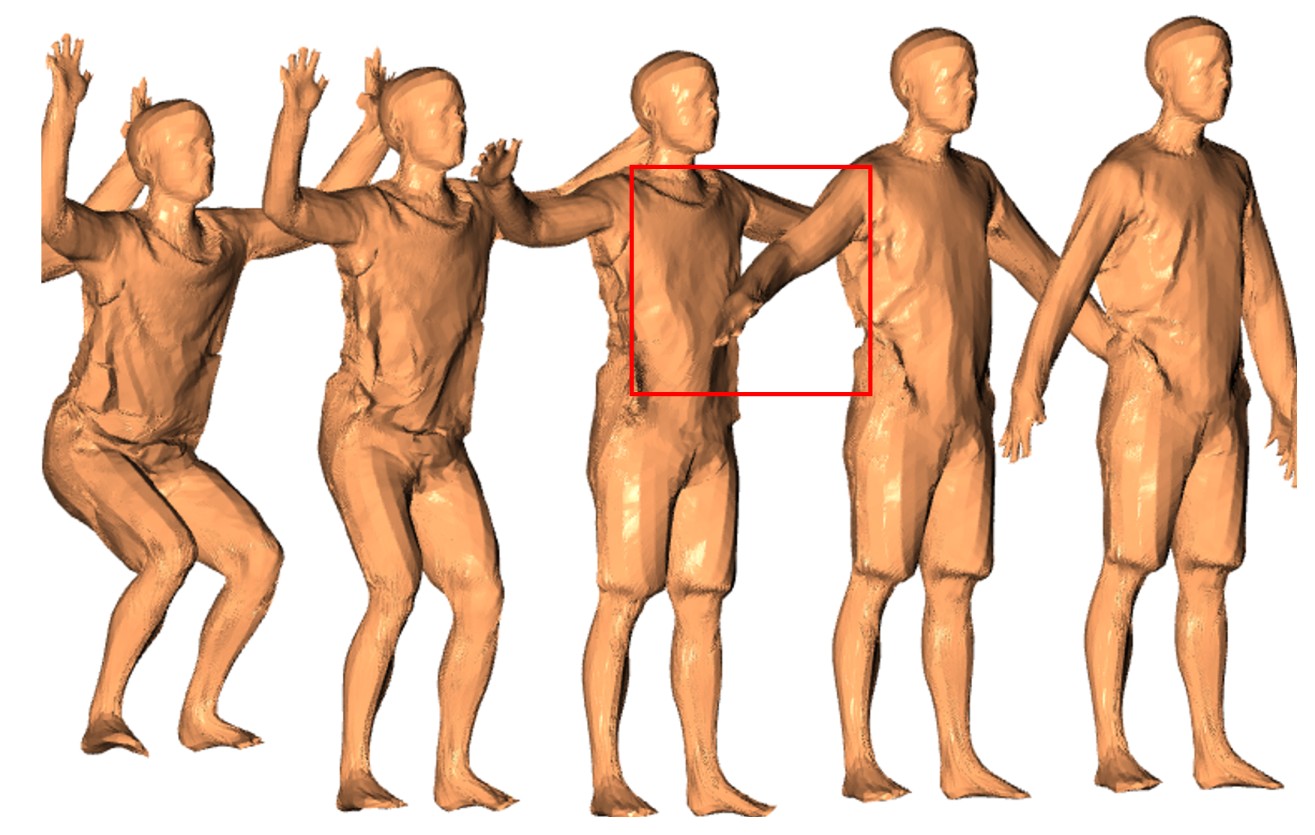}&
    \includegraphics[width=0.45\linewidth] {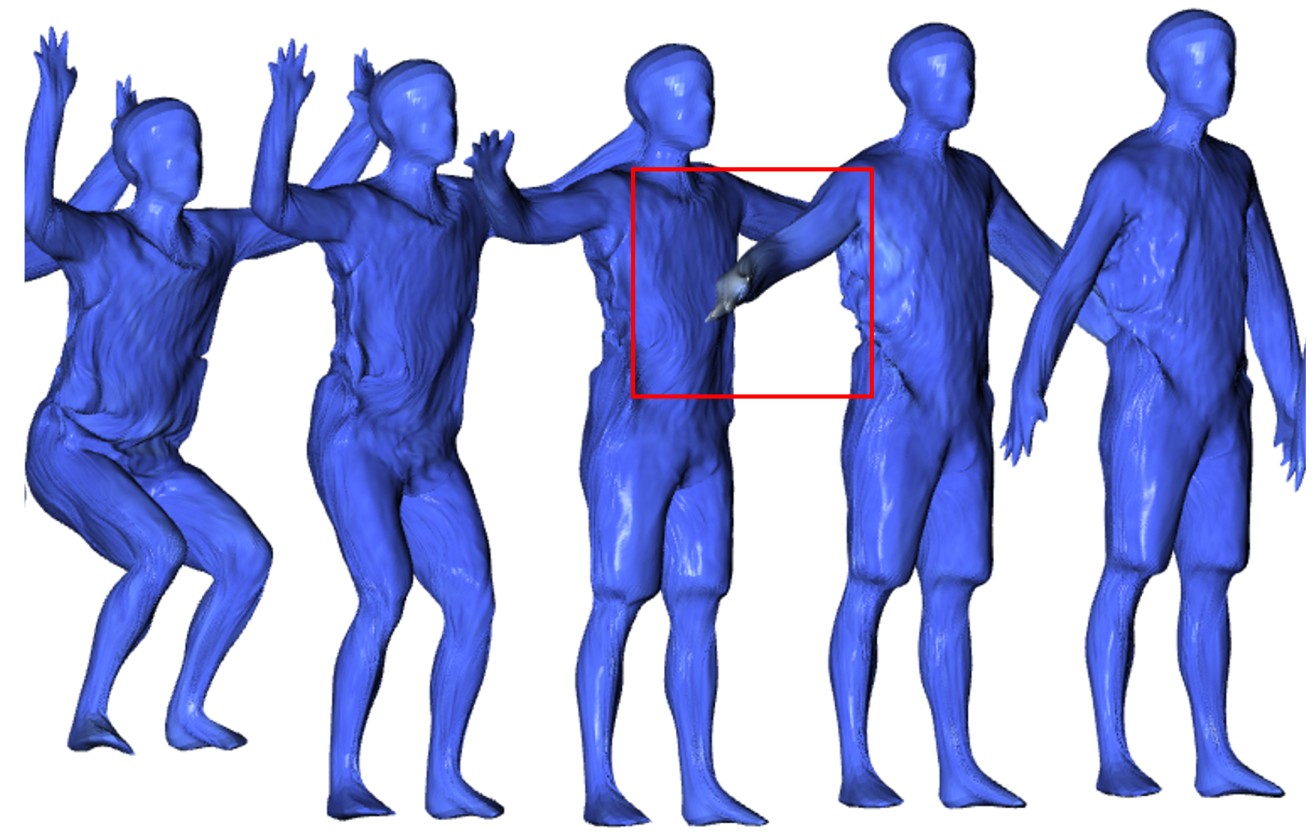} \\    
        (e) Ground truth.  &
        (f) Dynamic neural surface.
    \end{tabular}

   \caption{Example of the interpolation quality of D-SNS. We train D-SNS on a subset of $30$ temporal samples and visualize the time interval as a heatmap. Note that the D-SNS is able to faithfully represent even the detailed clothed human (from the CAPE dataset) and interpolate the missing sequences. The last row shows a zoom on the region highlighted in red. }
   \label{fig:resultsneuralrepresentationinterp}
\end{figure*}

\begin{figure*}[t]
  \centering
  \rule{\linewidth}{0.4pt}  
   \includegraphics[width=1\linewidth]{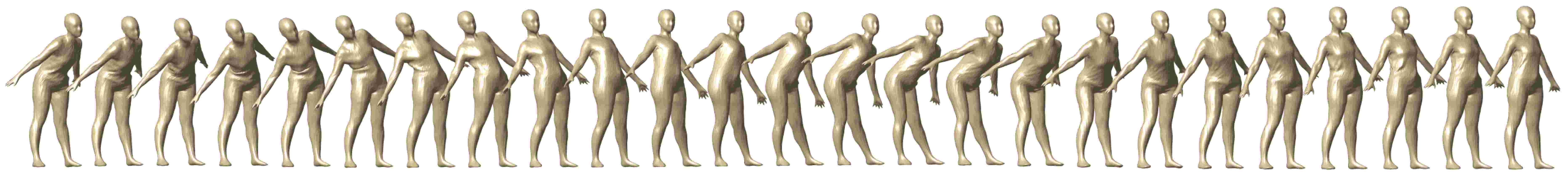}\\
   (a) Source.\\
   \includegraphics[width=1\linewidth]{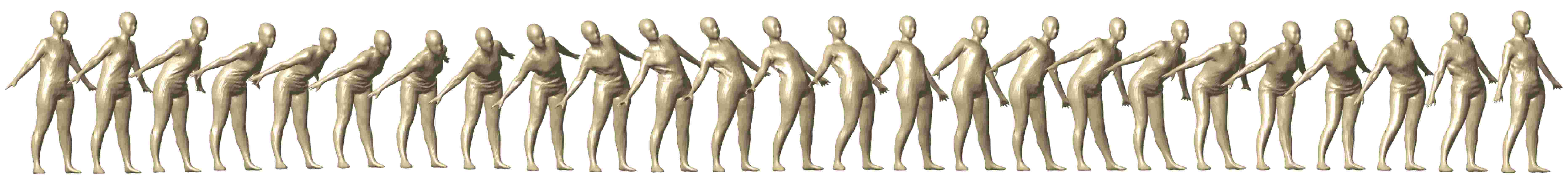}\\
   (b) Target before spatiotemporal registration.\\
    \includegraphics[width=1\linewidth]{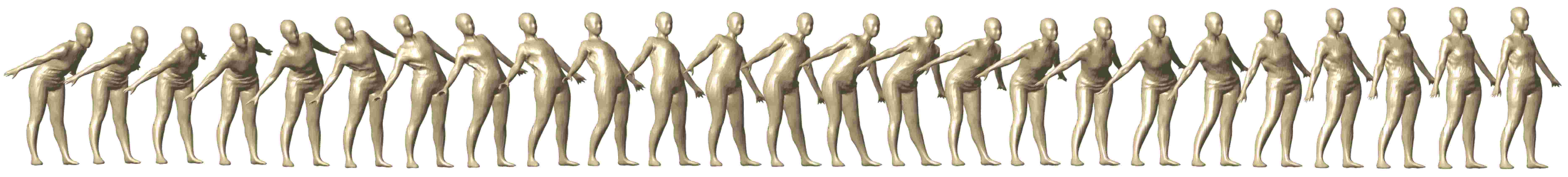}\\
   (c) Target after spatiotemporal registration.\\

  \rule{\linewidth}{0.4pt}  
  
   \includegraphics[width=1\linewidth]{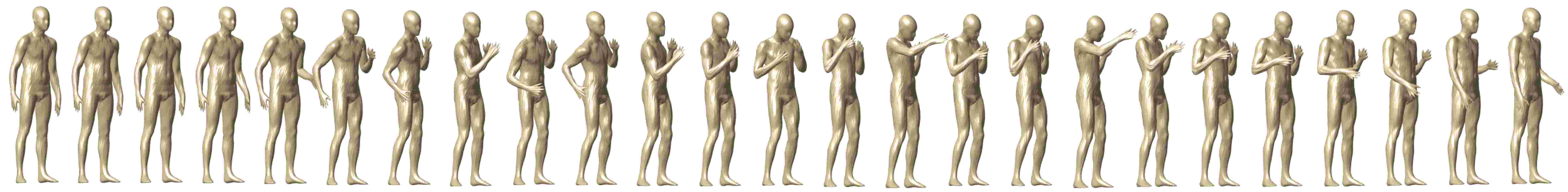}\\
   (a) Source.\\
   \includegraphics[width=1\linewidth]{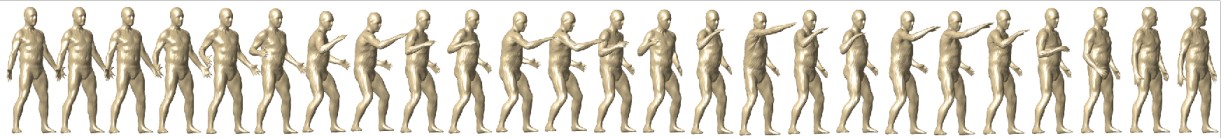}\\
   (b) Target before spatiotemporal registration.\\
    \includegraphics[width=1\linewidth]{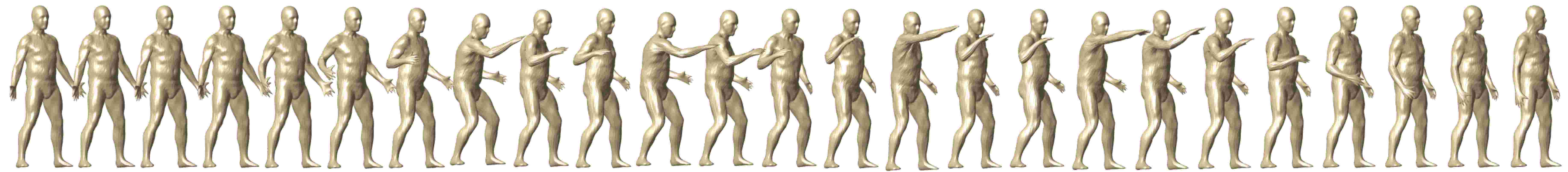}\\
   (c) Target after spatiotemporal registration.\\
   
  \rule{\linewidth}{0.4pt}
   \caption{Two examples of the spatiotemporal registration of  4D humans performing various actions. For each example, we show \textbf{(a)} the source 4D surface, \textbf{(b)} the target 4D surface before registration, and \textbf{(c)} the target 4D surface after registration using the proposed framework. Observe that the target neural surface after registration \textbf{(c)} is fully aligned with the source neural surface.
   }
   \label{fig:resultsspatiotemporalbody-suppl}
\end{figure*}

\begin{figure*}[t]
  \centering
   \rule{\linewidth}{0.4pt}
    \includegraphics[width=1\linewidth]{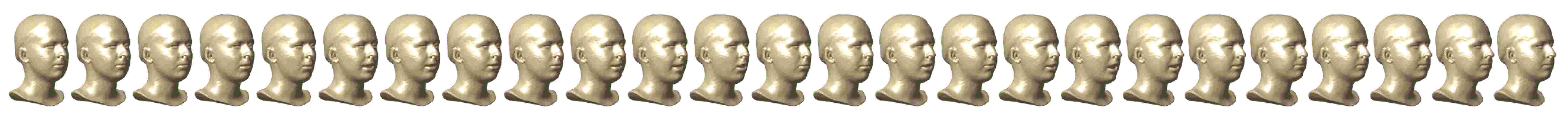}\\
   (a) Source.\\
   \includegraphics[width=1\linewidth]{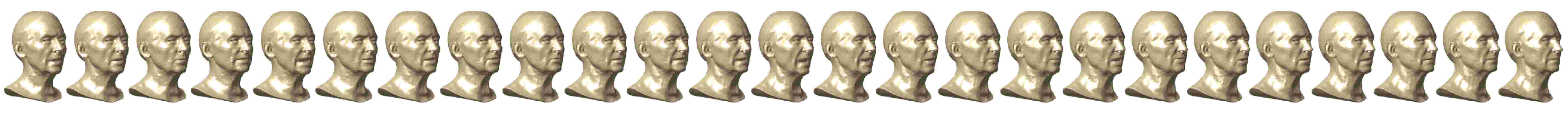}\\
   (b) Target before spatiotemporal registration.\\
    \includegraphics[width=1\linewidth]{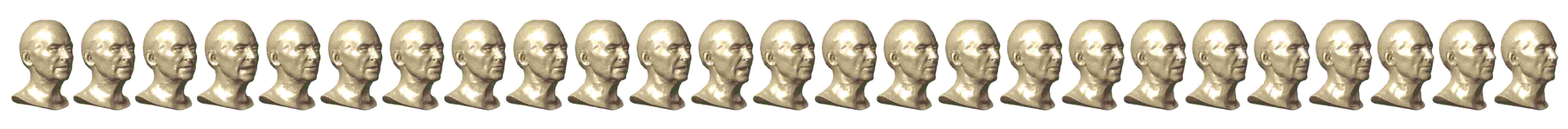}\\
   (c) Target after spatiotemporal registration.\\
   \rule{\linewidth}{0.4pt}

    \includegraphics[width=1\linewidth]{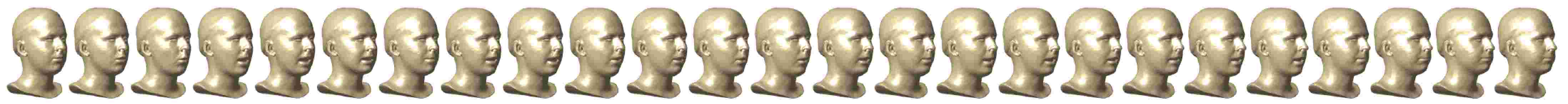}\\
   (a) Source.\\
   \includegraphics[width=1\linewidth]{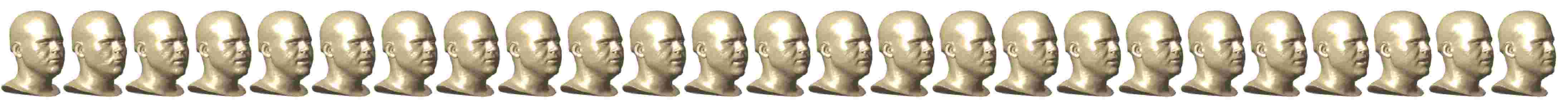}\\
   (b) Target before spatiotemporal registration.\\
    \includegraphics[width=1\linewidth]{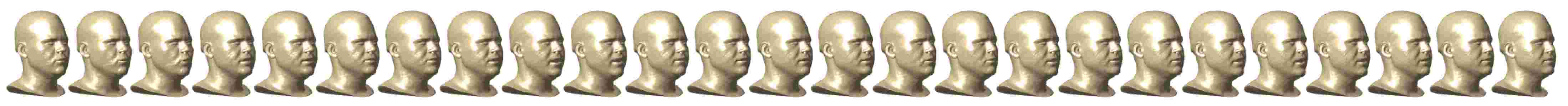}\\
   (c) Target after spatiotemporal registration.\\
   
   \rule{\linewidth}{0.4pt}

   \caption{Two examples of the spatiotemporal registration of  4D faces from the VOCA dataset. For each example, we show \textbf{(a)} the source 4D surface, \textbf{(b)} the target 4D surface before registration, and \textbf{(c)} the target 4D surface after registration using the proposed framework. Observe that the target neural surface after registration \textbf{(c)} is fully aligned with the source neural surface.}
   \label{fig:resultsspatiotemporalface-suppl}
\end{figure*}

\begin{table*}[t]
    \centering
        \begin{tabular}{@{}cccccc@{}}
            \toprule
             &         &        & \textbf{CAPE} &       &         \\
             \hline
             & squat  & chicken wings & twist tilt left        & punching & bend back and forth \\
            4D Atlas & 0.4607 & 1.1354 & 1.8219 & 1.404 & 1.6472  \\ 
            Ours & 0.1491 & 0.3199 & 0.2193  & 0.6064 & 0.4925\\ \hline
    
            &          &        & \textbf{DFAUST} &    &  \\
            \hline
             & punching & punching & jumping jacks & jumping jacks & punching \\
            4D Atlas & 1.7044 & 1.661 & 4.6092 & 1.8665 & 2.0516  \\ 
            Ours & 1.09 & 0.51 & 0.9023  & 0.637 & 0.721\\ \hline

            &         &        & \textbf{COMA}&   &      \\
            \hline
             & eyebrow & mouth extreme & high smile & lips back & mouth up \\
            4D Atlas & 0.0617 & 0.0493 & 0.0377 & 0.0626 & 0.0874  \\ 
            Ours & 0.0137 & 0.0173 & 0.0136  & 0.0296 & 0.0640\\ \hline

            &     &    &   \textbf{VOCA}&      &        \\
            \hline
             & sentence 3 & sentence 4 & sentence 1 & sentence 2 & sentence 3 \\
            4D Atlas & 0.6934 & 0.383 & 0.4494 & 0.4606 & 0.5266  \\ 
            Ours & 0.103 & 0.0691 & 0.0759  & 0.109 & 0.1400\\

            \bottomrule
        \end{tabular}
    \caption{Comparison of the proposed spatiotemporal registration with 4D Atlas~\cite{laga20234datlas}. The evaluation showcases the individual pair performance on all four datasets. Table~\ref{tab:temporalevaluationmain} provides the mean, standard deviation, and median for each dataset. }
    \label{tab:temporalevaluation-suppl}
\end{table*}

\begin{figure*}[ht]
    \centering
    \begin{tabular}{@{}c@{ }c@{ }c@{}}
        \includegraphics[width=0.335\linewidth]{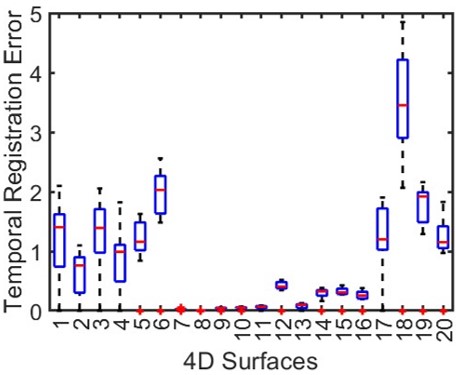} &
        \includegraphics[width=0.3\linewidth]{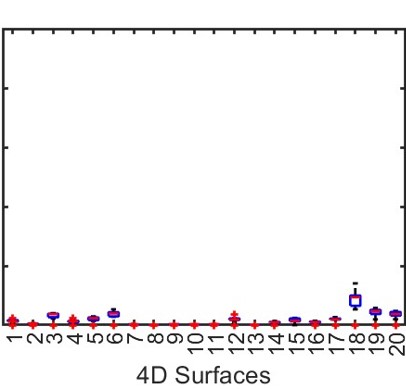} &
        \includegraphics[width=0.3\linewidth]{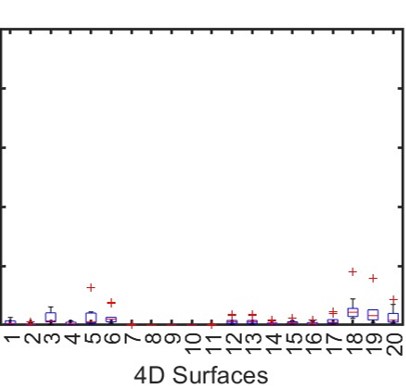} \\

        (a) Before registration. & (b) After registration 4D Atlas. & (c) After registration Ours.
    
    \end{tabular}
    \caption{Boxplot visualization of the spatiotemporal registration experiment performed in Figure~\ref{fig:evaluationboxplot} in the main manuscript. We  show the alignment error \textbf{(a)} before spatiotemporal registration, \textbf{(b)} after spatiotemporal registration using 4D Atlas~\cite{laga20234datlas}, and \textbf{(c)} after spatiotemporal registration using our framework.  
    }
    \label{fig:boxplot-suppl}
\end{figure*}

\begin{figure}[ht]
    \centering
    \includegraphics[width=0.75\linewidth]{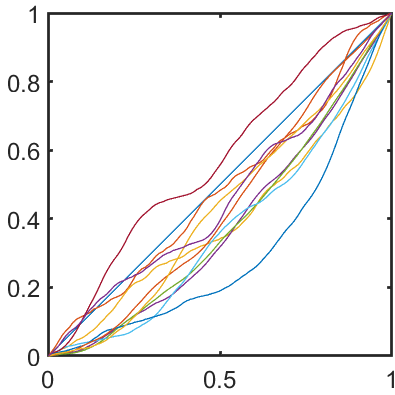}
    \caption{The $10$ temporal diffeomorphisms applied on the same 4D surfaces for the evaluation of our framework with 4D Atlas as shown in Figure~\ref{fig:evaluationboxplot}. Note that the range for each temporal diffeomorphism is from $[0,1] \to [0,1]$.}
    \label{fig:applieddiffeo-suppl}
\end{figure}

\begin{figure*}[t]
  \centering
    \includegraphics[width=0.97\linewidth]{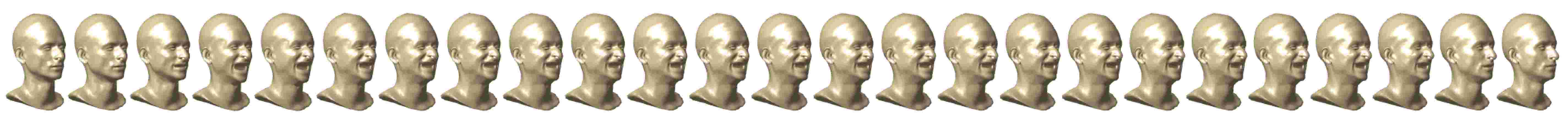}\\
    \includegraphics[width=0.97\linewidth]{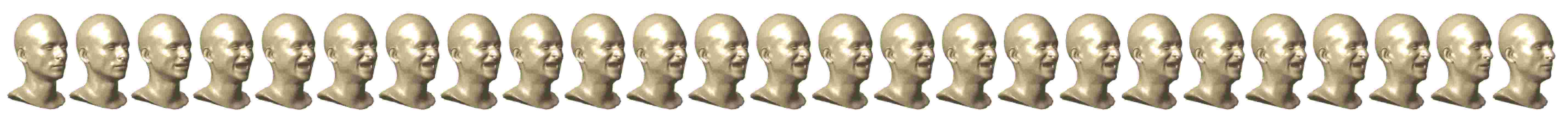}\\
    \includegraphics[width=0.97\linewidth]{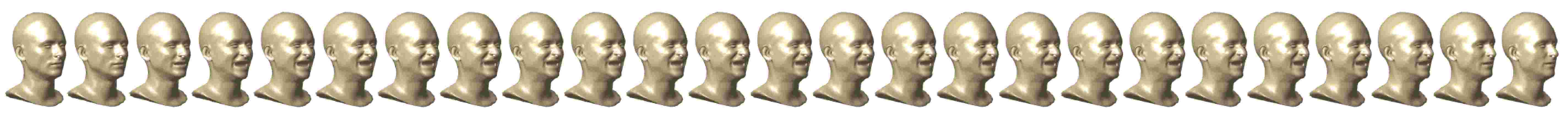}\\
    \fcolorbox{red}{white}{\includegraphics[width=0.97\linewidth]{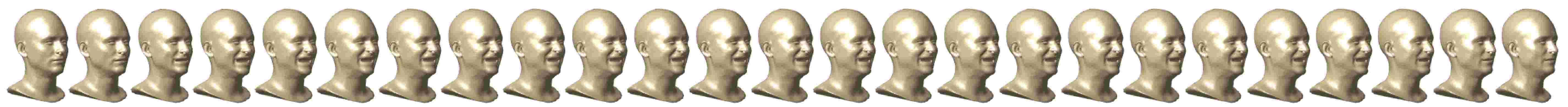}}\\
    \includegraphics[width=0.97\linewidth]{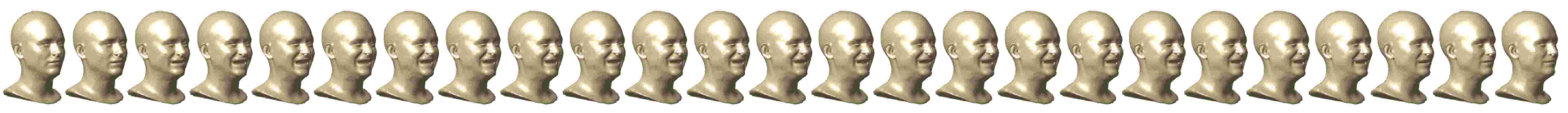}\\
    \includegraphics[width=0.97\linewidth]{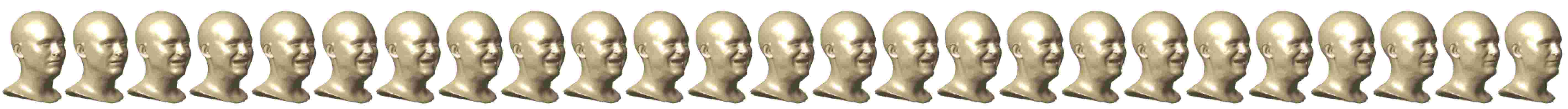}\\
    \includegraphics[width=0.97\linewidth]{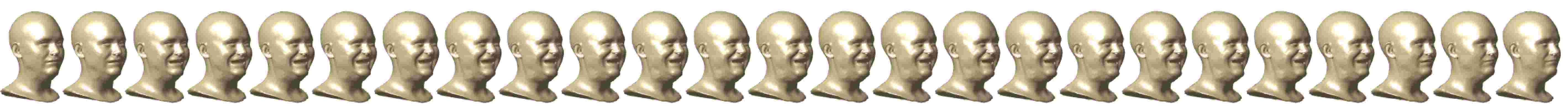}\\
    (a) Before spatiotemporal registration. 

    \includegraphics[width=0.97\linewidth]{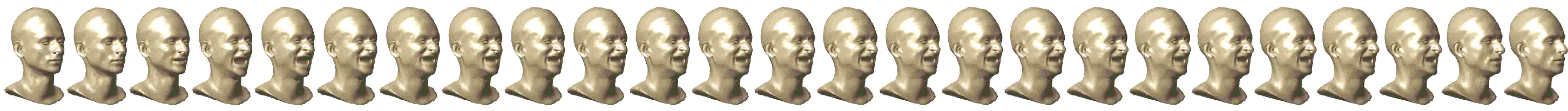}\\
    \includegraphics[width=0.97\linewidth]{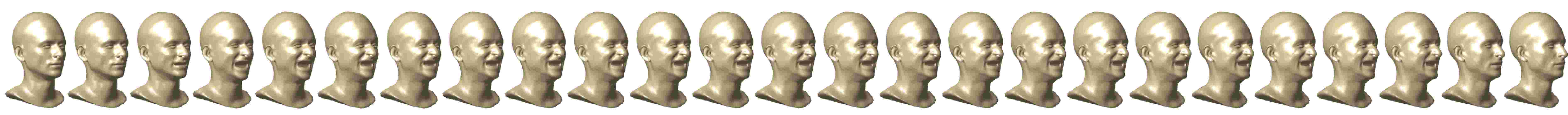}\\
    \includegraphics[width=0.97\linewidth]{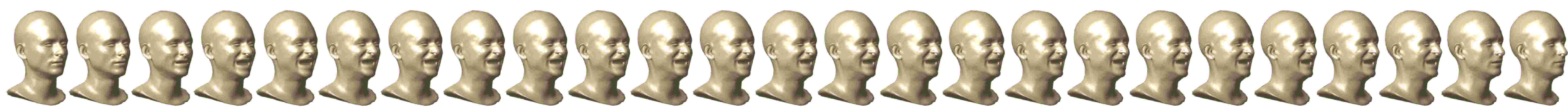}\\
    \fcolorbox{red}{white}{\includegraphics[width=0.97\linewidth]{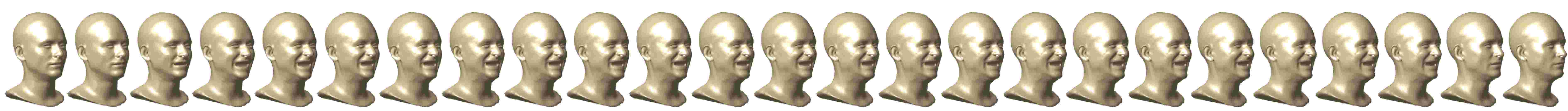}}\\
    \includegraphics[width=0.97\linewidth]{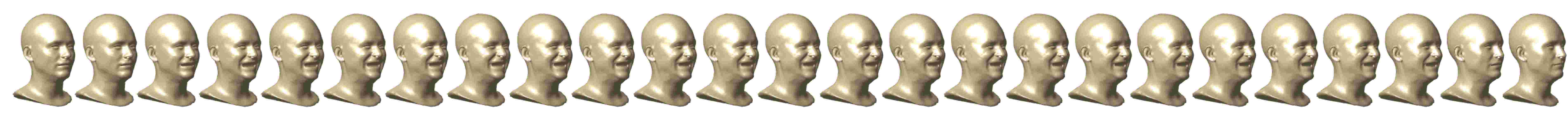}\\
    \includegraphics[width=0.97\linewidth]{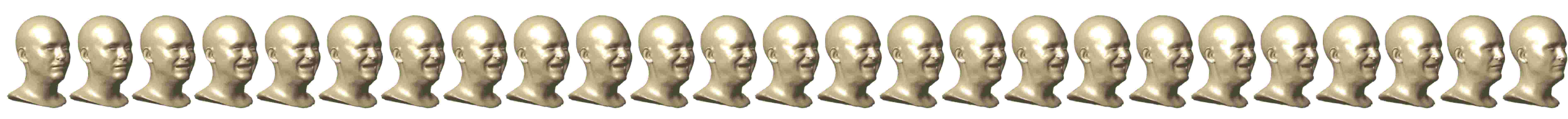}\\
    \includegraphics[width=0.97\linewidth]{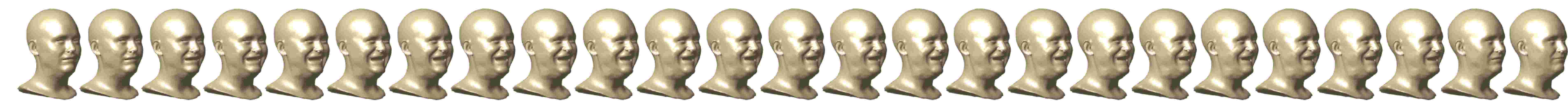}\\
    (b) After spatiotemporal registration.

   \caption{ 
   Example of a geodesic \textbf{(a)}  before and \textbf{(b)}  after registration between two 4D faces from the COMA dataset.  In each example, the first row corresponds to the source 4D surface, the last row corresponds to the target 4D surface, and the three intermediate rows correspond to intermediate 4D surfaces sampled at equidistance along the geodesic path between the source and target. Observe that before registration, the geodesics paths are not well-defined in the highlighted sequences. The highlighted row corresponds to the mean 4D surface.}
   \label{fig:results4dgeodesicsfaces-suppl}
\end{figure*}

\begin{figure*}[t]
  \centering
    \includegraphics[width=0.97\linewidth]{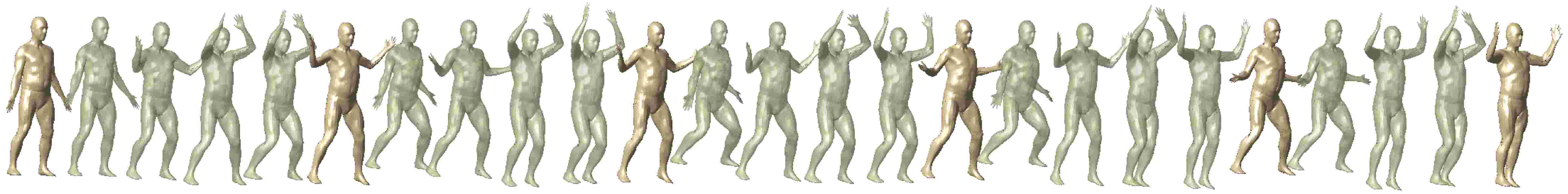}\\
    \includegraphics[width=0.97\linewidth]{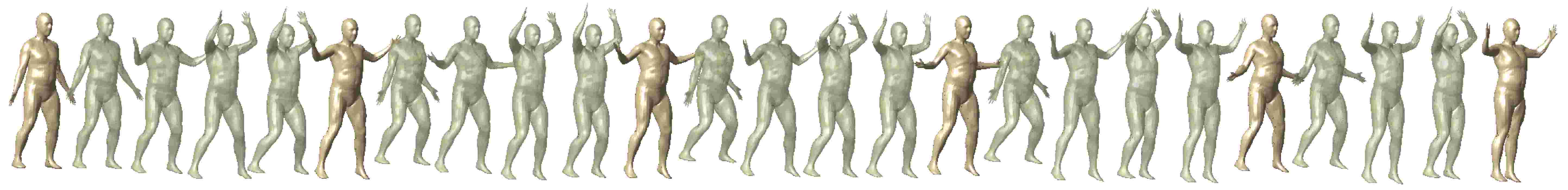}\\
    \fcolorbox{red}{white}{\includegraphics[width=0.97\linewidth]{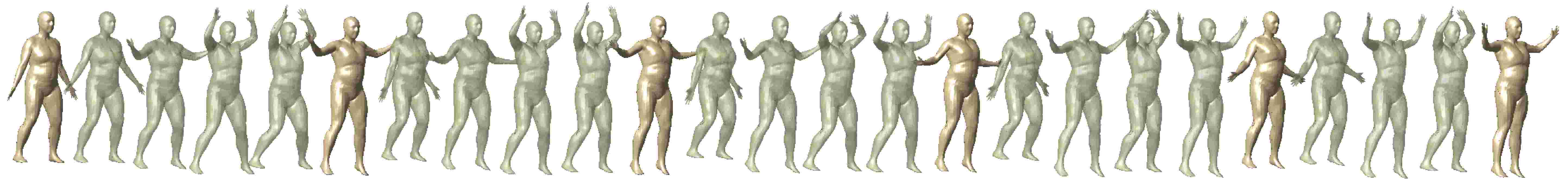}}\\
    \includegraphics[width=0.97\linewidth]{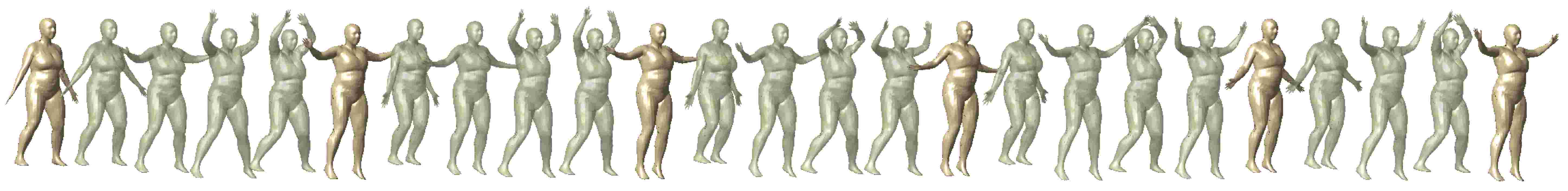}\\
    \includegraphics[width=0.97\linewidth]{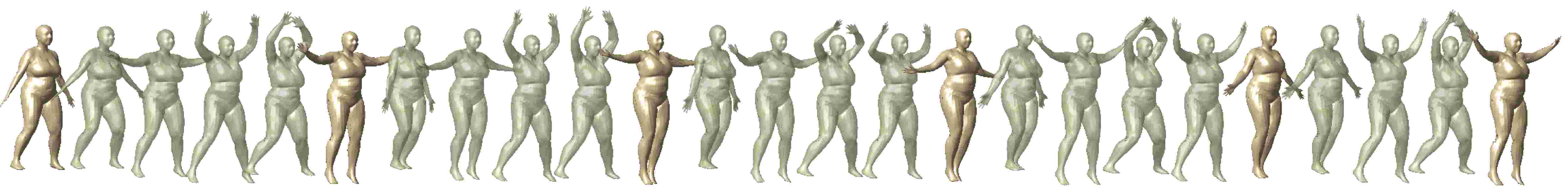}\\
    Before spatiotemporal registration. 


   \caption{Example of a geodesic before registration between two 4D surfaces performing a jumping action.  In this example, the first row corresponds to the source 4D surface, the last row corresponds to the target 4D surface, and the three intermediate rows correspond to intermediate 4D surfaces sampled at equidistance along the geodesic path between the source and target. Observe that the geodesics paths are not well-defined in the highlighted sequences. The highlighted row corresponds to the mean 4D surface.}
   \label{fig:results4dgeodesicshuman1-suppl}
\end{figure*}

\begin{figure*}[t]
    \centering
    \includegraphics[width=0.97\linewidth]{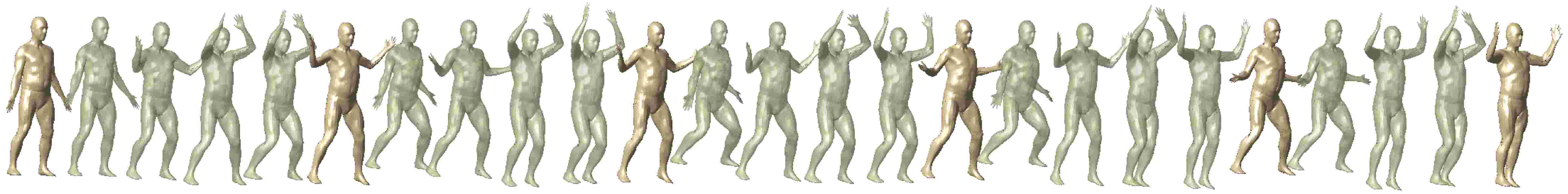}\\
    \includegraphics[width=0.97\linewidth]{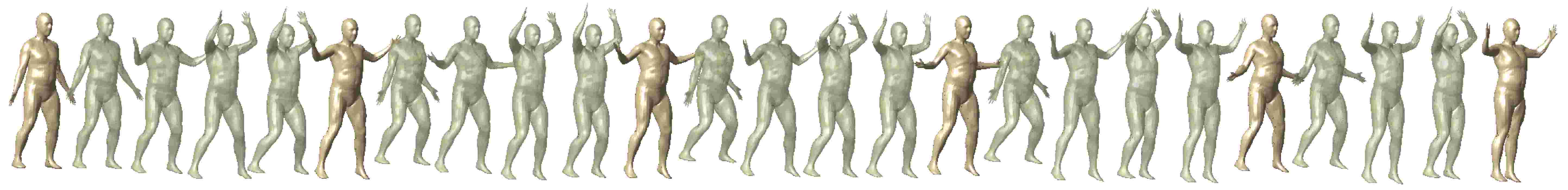}\\
    \fcolorbox{red}{white}{\includegraphics[width=0.97\linewidth]{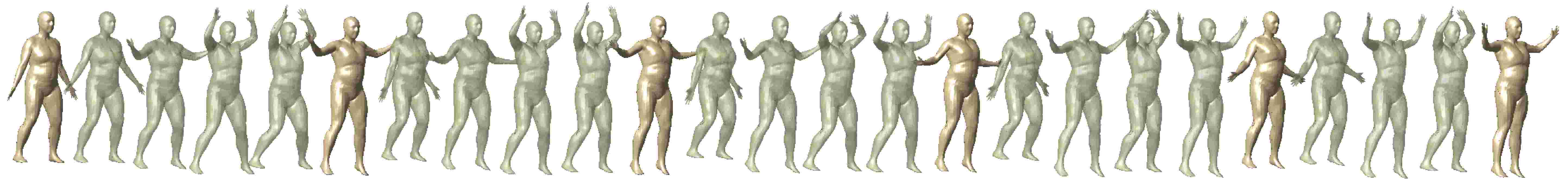}}\\
    \includegraphics[width=0.97\linewidth]{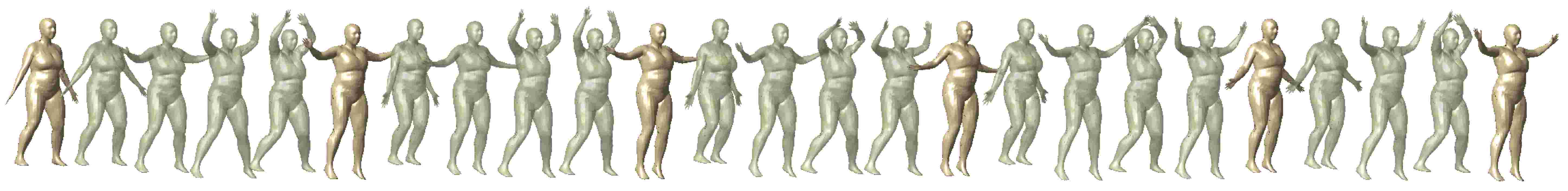}\\
    \includegraphics[width=0.97\linewidth]{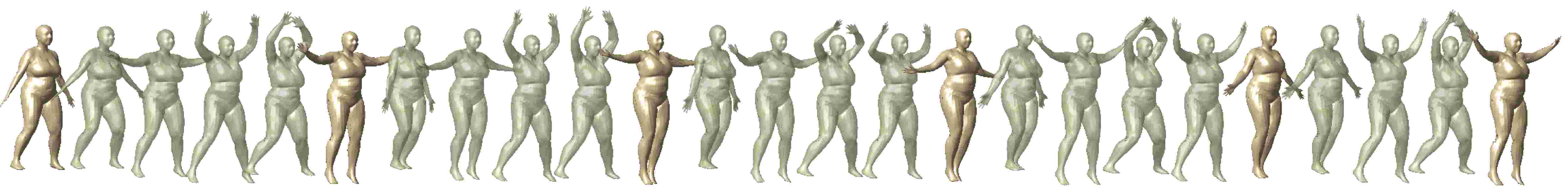}\\
    After spatiotemporal registration. 
    \caption{Example of a geodesic after registration between two 4D surfaces performing a jumping action.  In this example, the first row corresponds to the source 4D surface, the last row corresponds to the target 4D surface, and the three intermediate rows correspond to intermediate 4D surfaces sampled at equidistance along the geodesic path between the source and target. Observe that the geodesics paths are well-aligned in the highlighted sequences. The highlighted row corresponds to the mean 4D surface.}
   \label{fig:results4dgeodesicshuman2-suppl}
\end{figure*}

\begin{figure*}[t]
  \centering
    \includegraphics[width=0.97\linewidth]{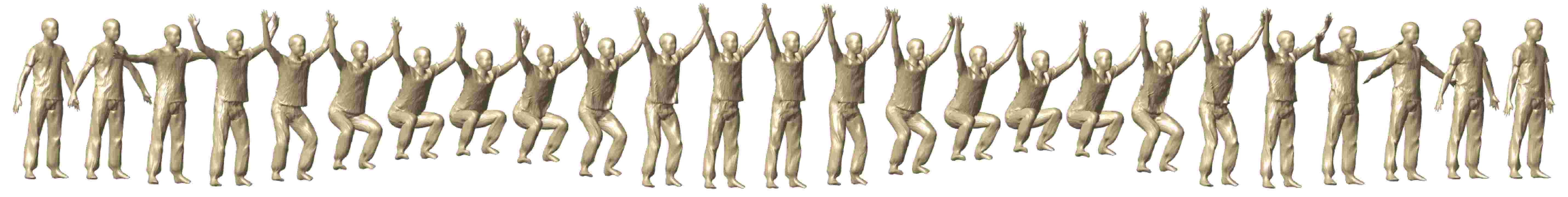}\\
    \includegraphics[width=0.97\linewidth]{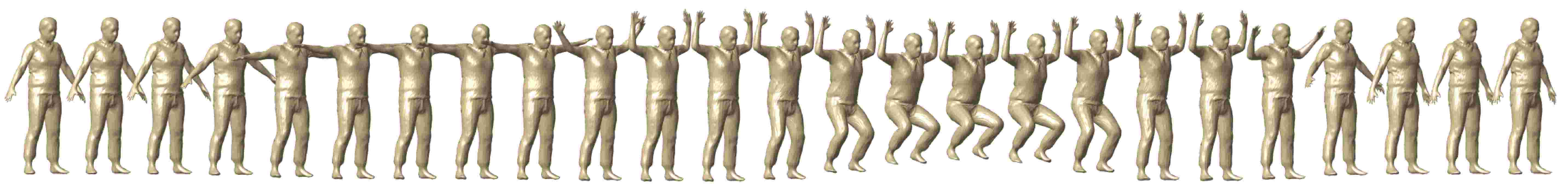}\\
    \includegraphics[width=0.97\linewidth]{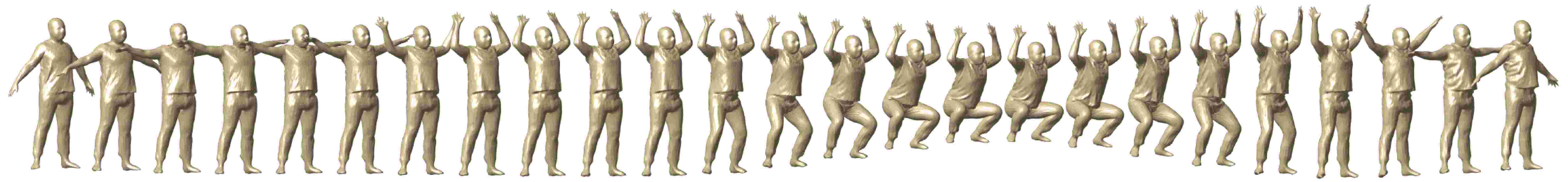}\\
    \includegraphics[width=0.97\linewidth]{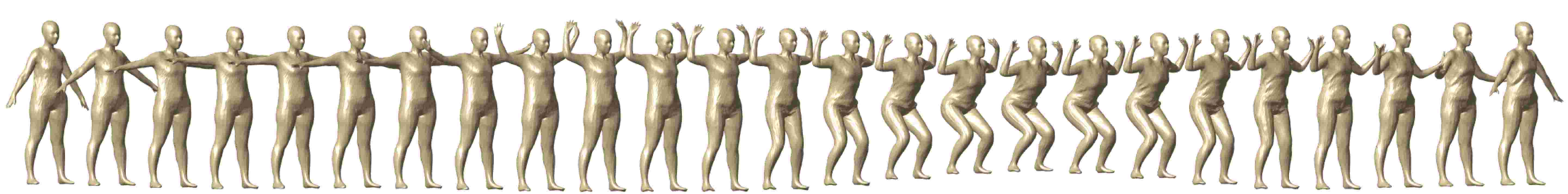}\\
    \includegraphics[width=0.97\linewidth]{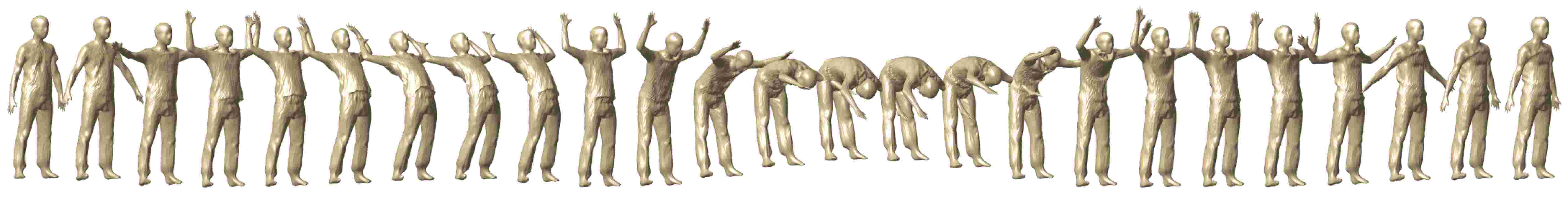}\\
    \includegraphics[width=0.97\linewidth]{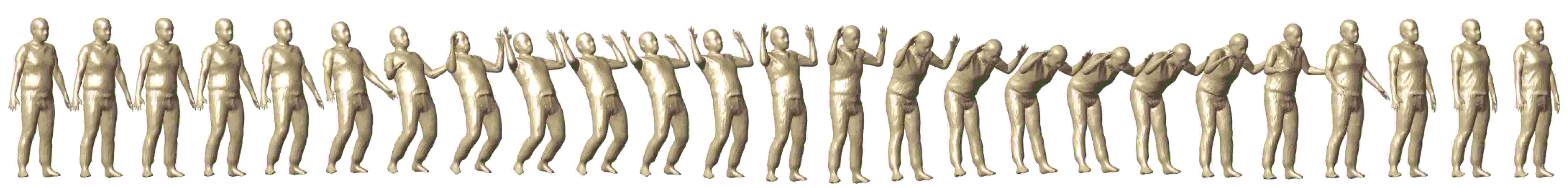}\\
    \fcolorbox{red}{white}{\includegraphics[width=0.97\linewidth]{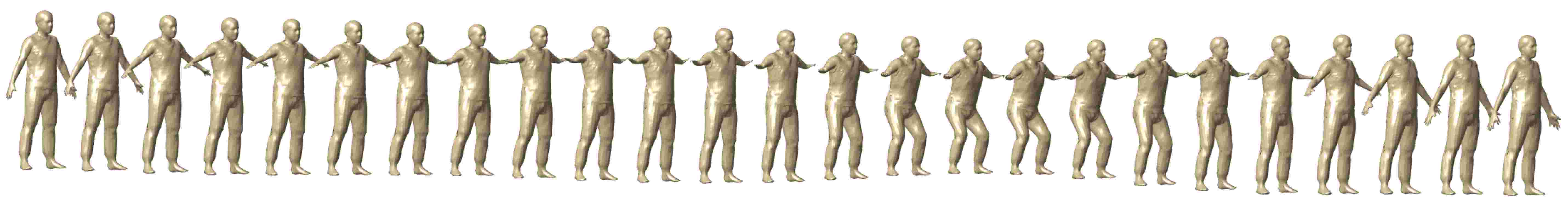}}\\

   \caption{Example of a mean 4D surface (highlighted in red) computed on six 4D surfaces from  the CAPE dataset \textbf{before their spatiotemporal registration}. The input 4D surfaces perform different actions:  the 4D surfaces in the first four rows perform a squat action, while the last two perform a back and forward bending action. Observe how misaligned the 4D surfaces are before their co-registration and mean computation. 
   }
   \label{fig:results4dmeansurfaces-human1-suppl}
\end{figure*}

\begin{figure*}[t]
  \centering

    \includegraphics[width=0.97\linewidth]{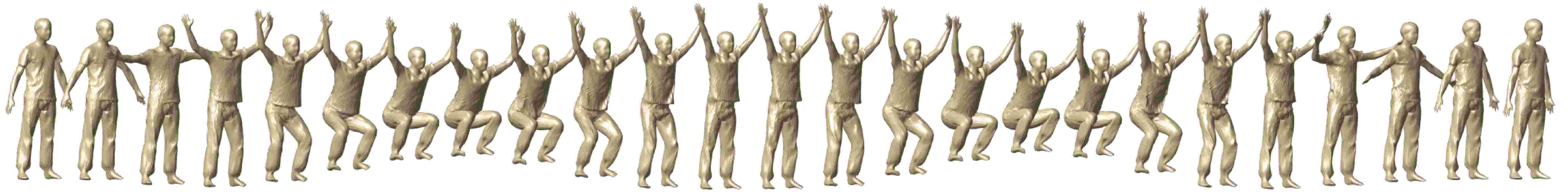}\\
    \includegraphics[width=0.97\linewidth]{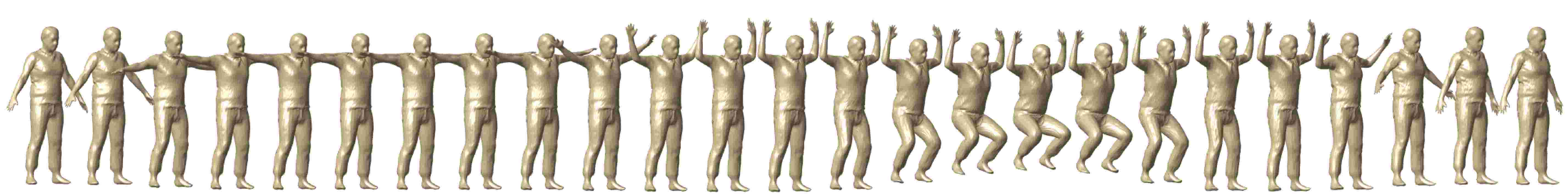}\\
    \includegraphics[width=0.97\linewidth]{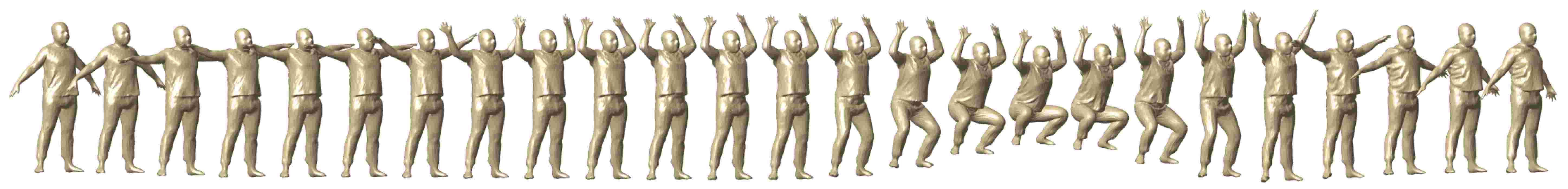}\\
    \includegraphics[width=0.97\linewidth]{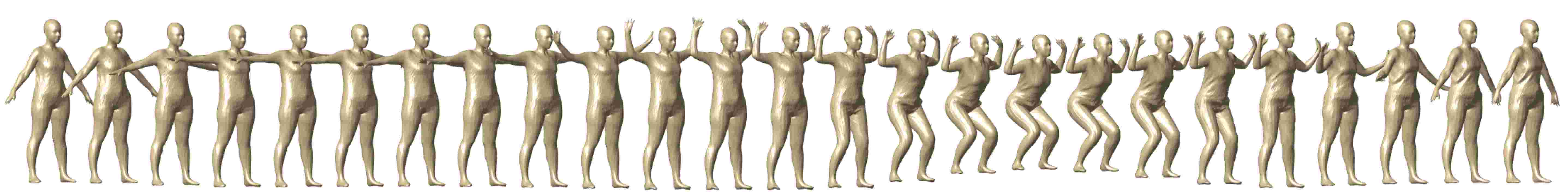}\\
    \includegraphics[width=0.97\linewidth]{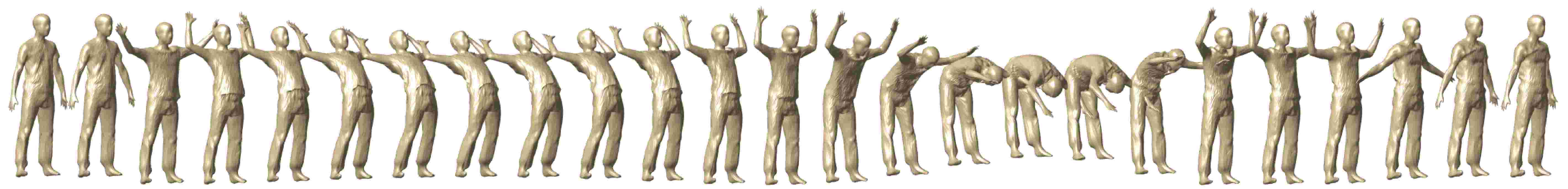}\\
    \includegraphics[width=0.97\linewidth]{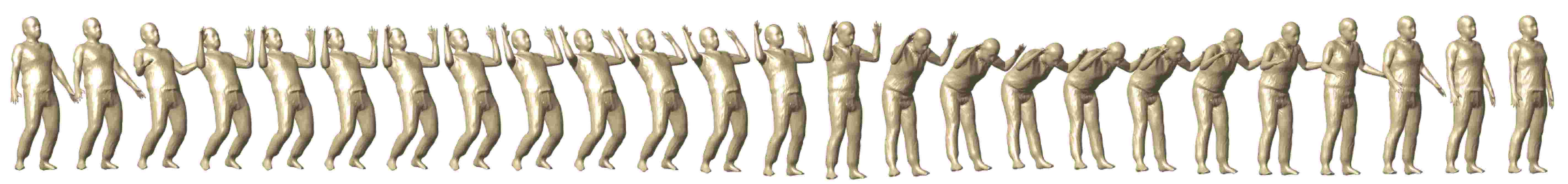}\\
    \fcolorbox{red}{white}{\includegraphics[width=0.97\linewidth]{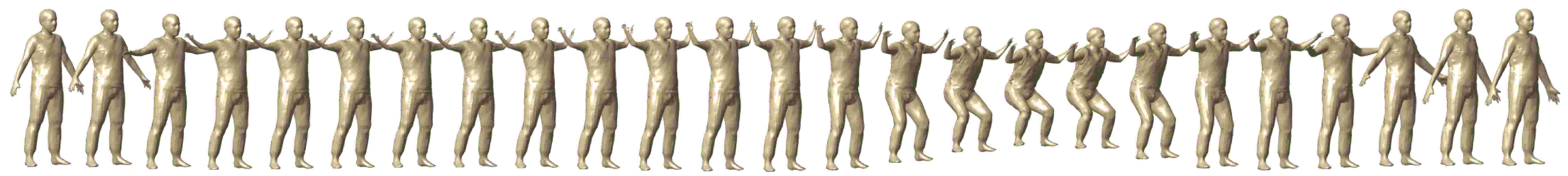}}\\

   \caption{Example of a mean 4D surface (highlighted in red) computed on six 4D surfaces from the CAPE dataset \textbf{after their spatiotemporal registration}. The input 4D surfaces perform different actions:  the 4D surfaces in the first four rows perform a squat action while the last two perform a back and forward bending action. Observe how aligned the 4D surfaces become after their co-registration and mean computation compared to Figure~\ref{fig:results4dmeansurfaces-human1-suppl}. 
   }
   \label{fig:results4dmeansurfaces-human2-suppl}
\end{figure*}

\begin{figure*}[t]
  \centering
    \includegraphics[width=0.97\linewidth]{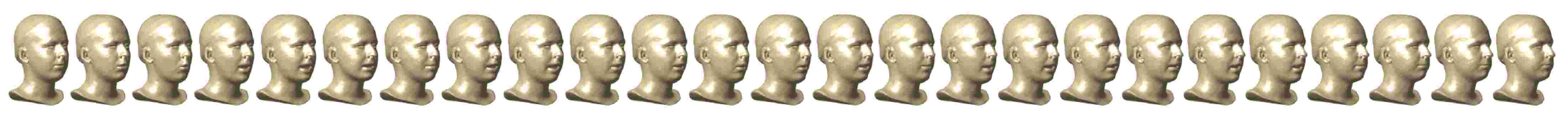}\\
    \includegraphics[width=0.97\linewidth]{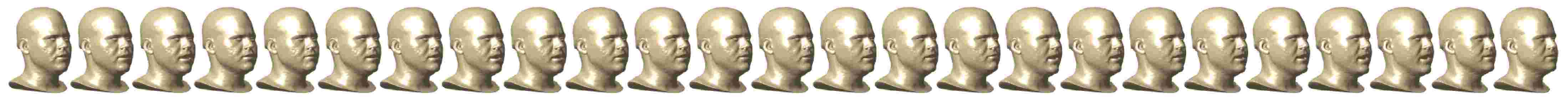}\\
    \includegraphics[width=0.97\linewidth]{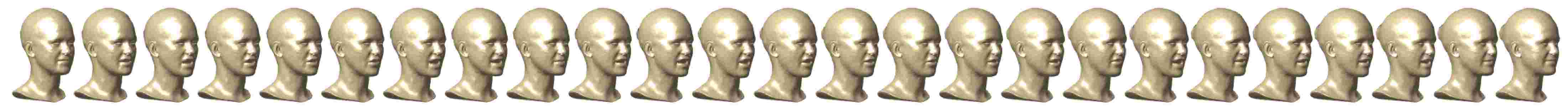}\\
    \includegraphics[width=0.97\linewidth]{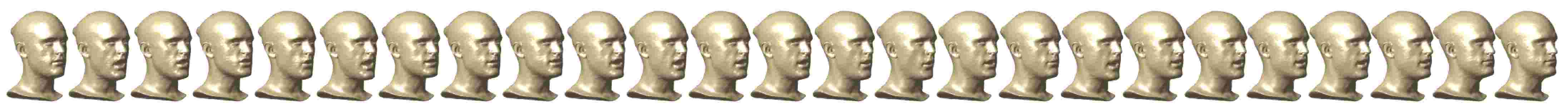}\\
    \includegraphics[width=0.97\linewidth]{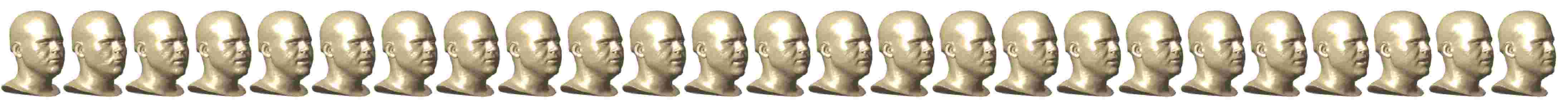}\\
    \includegraphics[width=0.97\linewidth]{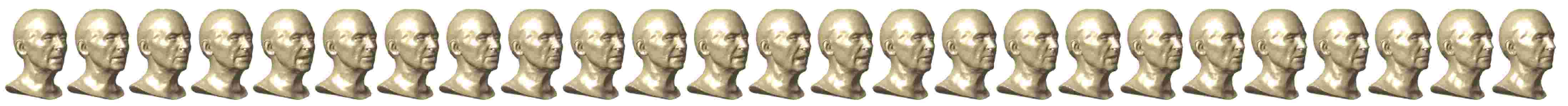}\\
    \fcolorbox{red}{white}{\includegraphics[width=0.97\linewidth]{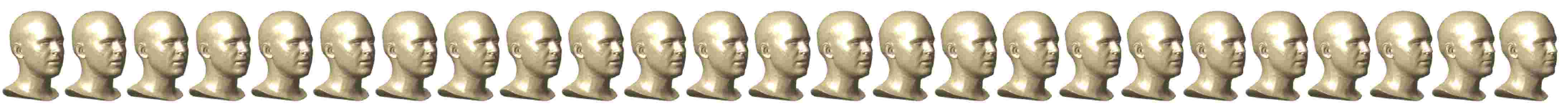}}\\
    (a) Before spatiotemporal registration.

    \includegraphics[width=0.97\linewidth]{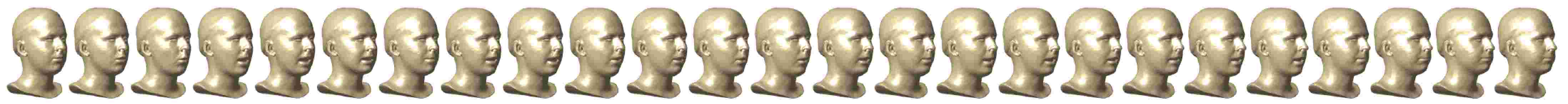}\\
    \includegraphics[width=0.97\linewidth]{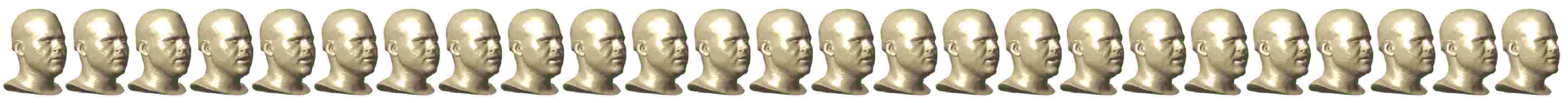}\\
    \includegraphics[width=0.97\linewidth]{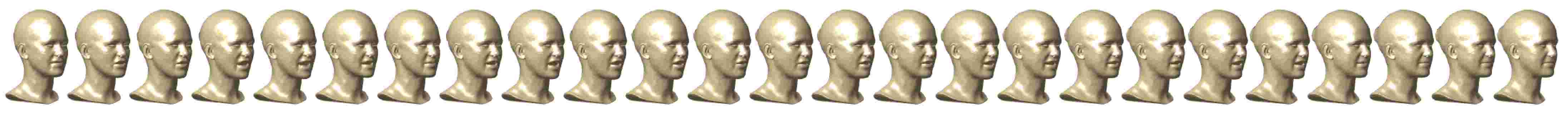}\\
    \includegraphics[width=0.97\linewidth]{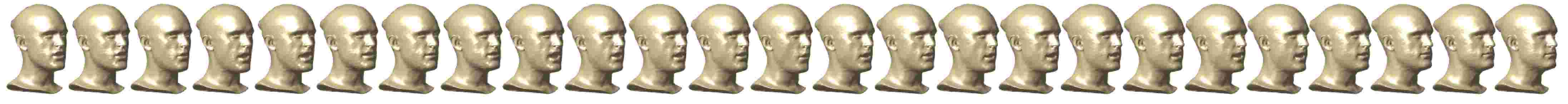}\\
    \includegraphics[width=0.97\linewidth]{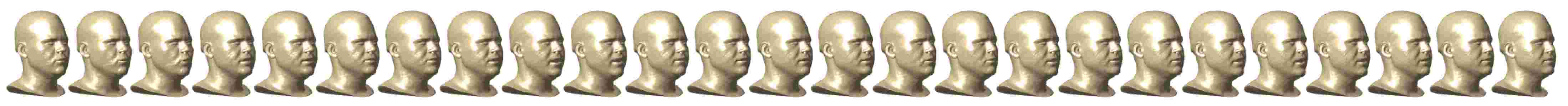}\\
    \includegraphics[width=0.97\linewidth]{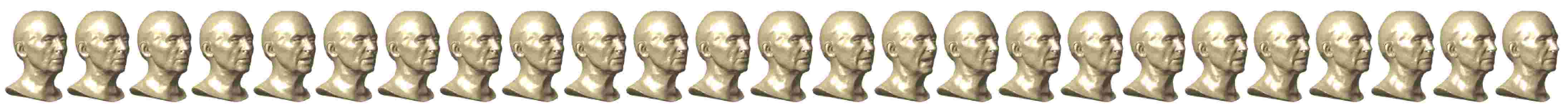}\\
    \fcolorbox{red}{white}{\includegraphics[width=0.97\linewidth]{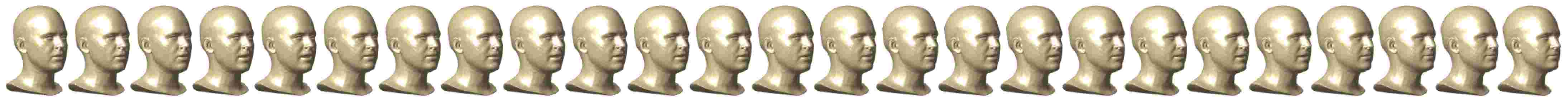}}\\
    (b) After spatiotemporal registration. 

   \caption{Example of a mean 4D surface (highlighted in red) computed on six 4D surfaces  from the VOCA dataset \textbf{(a)} before and \textbf{(b)}  after their spatiotemporal registration. The input 4D surfaces speak different sentences:  the 4D surfaces in the first four rows speak the same sentence while the last two rows speak a different sentence. Observe how aligned the 4D surfaces become after their co-registration and mean computation. 
   }
   \label{fig:results4dmeansurfaces4-faces-suppl}
\end{figure*}

\section{Implementation details}
\label{suppl:implem}

\subsection{D-SNS network}
We employ a Multi-Layer Perceptron (MLP) composed of six residual blocks. Each block consists of two layers of $1024$ neurons each. We use positional encoding of both space and time. The output layer of each block uses  SoftPlus as activation function to represent smooth and continuous 4D surfaces.  Figure~\ref{fig:network-design} summarizes the detailed architecture. 

\vspace{6 pt}
\noi\textbf{Training.} We learn a continuous representation $\spatiotemporalsurface$ of a discrete 4D surface using D-SNS. We first spherically parameterized the 4D surfaces, which consist of a set of triangular meshes, with the approach of~\cite{kurtek2013landmark}. We then map the mesh sequences to a temporal domain and allocate a time value in the range of $[0,1]$. Next, we train D-SNS for this discrete 4D surface by defining a batch size of $80,000$ surface points, which are randomly selected.  For each point $\pointonsurface$ we have its associated point on sphere $\pointonsphere$ and a time instance $\thetime$. We then parse this batch to the D-SNS network, which outputs the predicted points on the surface $\surfacepoint^*$. We minimize the  $\ltwo$ loss between the D-SNS represented points $\surfacepoint^*$ and the discrete points $\pointonsurface$. We have noticed that using the random sampling of surface points from the meshes helps our D-SNS network to converge faster. It also results in smooth and continuous 4D surfaces.


\subsection{Spatial diffeomorphism}
As discussed in Section~\ref{sec:spatial_registration}, we select 3D instances $\surfaceone, \surfacetwo$ from their 4D surfaces $\spatiotemporalsurface_1, \spatiotemporalsurface_2$. We then find the optimal rotation $\rotation \in \rotations$ and diffeomorsphism $\diffeo \in \diffeos$ such that when $\rotation (\surfacetwo \circ \diffeo)$ is  spatially register $\surfacetwo$ onto $\surfaceone$.

We use a gradient descent-based optimization method that finds the optimal diffeomorphism $\diffeo^*$ and rotation $\rotation^*$  in the SRNF space. As discussed in our approach, we apply the spatial registration framework directly to the neural functions. We  keep optimizing for diffeomorphism $\diffeo$ using a weighted sum of spherical harmonic basis and rotation $\rotation$ using Singular Value Decomposition (SVD). We apply these on the unit sphere and parse the reparameterized unit sphere to the D-SNS network, which results in a spatially registered neural function $\surfacetwo$ that is as close as possible to $\surfaceone$.

\subsection{Time warping network}
The time-warping network is an MLP that finds the optimal temporal alignment between the SRVFs $\srvfone, \srvftwo$ of two curves $\pcacurveone, \pcacurvetwo$ obtained from spatially registered D-SNS $\spatiotemporalsurface_1, \spatiotemporalsurface_2$; see Figure~\ref{fig:network-design} for the detailed architecture.

\vspace{6 pt}
\noi\textbf{Training.} We obtain the 4D surfaces $\spatiotemporalsurface_1$ and $\spatiotemporalsurface_2$ at a spherical resolution of $32 \times 32$ with 50 time samples $\thetime\}_{i=1}^{50}$. First, using PCA, we map the surfaces $\spatiotemporalsurface_1$ and $\spatiotemporalsurface_2$ to a low dimensional space to obtain two curves $\pcacurveone=\pcamap(\spatiotemporalsurface_1), \pcacurvetwo=\pcamap(\spatiotemporalsurface_2)$. We then compute their SRVF $\srvfone = \srvfmap(\pcacurveone), \srvftwo = \srvfmap(\pcacurvetwo)$. Since the $\ltwo$ metric in the SRVF space measures nonrigid deformations of curves in the original space, we train the time warping network using the $\ltwo$ loss, \ie $\timewarp^* = \parallel \srvfone - \srvftwo \circ \timewarp \parallel$. To ensure that $\timewarp$ is a diffeomorphism, it needs to be a monotonically increasing function on the temporal domain $[0,1]$. To enforce this, we apply a regularization term that enforces the  first derivative of the network with respect to time $t$ to be non-negative. We also apply  the Sigmoid activation function to the output of the time-warping  network to keep its output within the bounds of $[0,1]$.

We initialize the training with the parameters of a pre-trained time-warping network that is overfitted, in an offline pre-processing step, to the identity diffeomorphism. We then refine the training for $2,000$ epochs and continuously change the timestamps after every $200$ epoch. This way, the time-warping network is able to learn a continuous temporal representation that aligns $\spatiotemporalsurface_2$ to $\spatiotemporalsurface_1$.

\section{Datasets}
\label{suppl:dataset}
 
We have evaluated the proposed framework on:
\begin{itemize}
    \item MPI DFAUST~\cite{dfaust:CVPR:2017}, which contains high-resolution 4D body scans of $10$ human subjects in motion, captured at $60$ fps; 
    
    \item  VOCA~\cite{VOCA2019}, which contains high-resolution 4D facial scans of $12$ subjects speaking various sentences; 
    
    \item  MPI COMA~\cite{COMA:ECCV18}, which contains 4D facial scans of $12$ subjects performing various facial expressions; and 
    
    \item  MPI 4D CAPE~\cite{CAPE:CVPR:20}, which contains high-resolution 4D full-body scans of $10$ male and $5$ females in clothing. 
\end{itemize}

\noi The datasets come with registered triangular meshes. We spherically parameterize these datasets using the implementation~\cite{kurtek2013landmark} of  Praun and Hoppe's approach~\cite{praun2003spherical}. We then generate random diffeomorphisms to simulate non-registered surfaces.

\section{Results}
\label{suppl:results}

In this section, we show additional results of our neural framework that could not fit within the page limit of the main manuscript. It also reproduces the figures of the main manuscript in high resolution.

\subsection{Dynamic Spherical Neural Surfaces}
\label{supplsec:dsns-results}

Figure~\ref{fig:resultsneuralrepresentation1-suppl} provides more quantitative results on the four datasets. Similar to  Figure~\ref{fig:resultsneuralrepresentation1} in main manuscript, here we measure the representation capability of the proposed neural representation using the pointwise error between the neural surfaces and the original ground-truth surfaces. As one can see in Figure~\ref{fig:resultsneuralrepresentation1-suppl}, the error is smaller than $0.01\%$. Note that all the surfaces have been normalized for scale to fit within a unit sphere centered at the origin.

Figure~\ref{fig:resultsneuralrepresentationinterp}, on the other hand, demonstrates the interpolation ability of our representation; see  Table~\ref{tab:neuralrepresentationinterp} in the main manuscript for a quantitative evaluation. In this experiment, the neural representation was trained only on $30$ temporal samples of the entire sequences. Yet, the method is able to interpolate the missing frames and generate a plausibly smooth 4D surface. For example,  the clothed 4D human from the CAPE dataset with high clothing wrinkles is accurately represented and faithfully interpolated using the proposed D-SNS representation.

\subsection{Spatiotemporal registration}

Figures~\ref{fig:resultsspatiotemporalbody-suppl} and~\ref{fig:resultsspatiotemporalface-suppl} show examples of the spatiotemporal registration of pairs of 4D surfaces. In Figure~\ref{fig:resultsspatiotemporalbody-suppl}, we show two examples of   4D humans before and after their temporal registration. Figure~\ref{fig:resultsspatiotemporalface-suppl}  shows two examples of 4D faces before and after their temporal registration. These examples demonstrate  that our neural framework is able to spatiotemporally register complex body articulations and facial expressions.

Figure~\ref{fig:boxplot-suppl} shows the quantitative evaluation of the temporal registration on the same 4D surfaces as the ones shown in Figure~\ref{fig:evaluationboxplot} in the main manuscript. In this experiment, we use the evaluation framework proposed in 4D Atlas~\cite{laga20234datlas}.  Note that, we have changed the range of the  $Y$ axis from $0-1$ to $0-5$ to faithfully represent the error before  and  after registration.

Figure~\ref{fig:applieddiffeo-suppl}, on the other hand, shows the $10$ temporal diffeomorphisms applied to perturb the same 4D surfaces  for quantitative evaluation performed in Figure~\ref{fig:evaluationboxplot} of the main manuscript.

Table~\ref{tab:temporalevaluation-suppl} expands the results of Table~\ref{tab:temporalevaluationmain} in the main manuscript by providing the error on each individual 4D surface.   In this experiment, we measure the geodesic distance between the registered 4D surfaces using our method and 4D Atlas method. Note that the smaller the geodesic distance is, the better is the alignment.

\subsection{4D geodesics}
Figure~\ref{fig:results4dgeodesicsfaces-suppl} shows an example of a geodesic of 4D faces from COMA dataset.
Figure~\ref{fig:results4dgeodesicshuman1-suppl} and Figure~\ref{fig:results4dgeodesicshuman2-suppl}, on the other hand, show a high-resolution version of the example of Figure~\ref{fig:results4dgeodesicshuman} in the main manuscript. 
Figure~\ref{fig:results4dgeodesicshuman1-suppl} shows the 4D surfaces before registration; notice how misaligned is the mean 4D surface  (highlighted in red) with the input surfaces.  Figure~\ref{fig:results4dgeodesicshuman2-suppl} shows the same geodesic after spatiotemporal registration of the source and target 4D surfaces.

\subsection{Co-registration and mean 4D surfaces}
Figure~\ref{fig:results4dmeansurfaces-human1-suppl} and Figure~\ref{fig:results4dmeansurfaces-human2-suppl} shows an example of the 4D mean surface of a set of 4D neural surfaces, from the CAPE dataset. In this document, we show the same example before registration (Figure~\ref{fig:results4dmeansurfaces-human1-suppl}) and after registration (Figure~\ref{fig:results4dmeansurfaces-human2-suppl}). The 4D surfaces in these two figures  perform a squat action, and the following two rows perform a bending action. Note that squat action is repetitive, and the number of cycles differs from one 4D surface to another. In particular, the 4D surface in the first row performs two squats while the remaining 4D surfaces perform a single one. Despite this complexity, the proposed approach is able to co-register the 4D surfaces and compute a plausible 4D mean that is as close as possible to all the other 4D surfaces.

Similarly, Figure~\ref{fig:results4dmeansurfaces4-faces-suppl} shows the co-registration and  4D mean surface of six 4D faces, from the VOCA dataset, speaking sentences. In this example, the first four rows speak a different sentence than the last two rows. The  facial expressions on each 4D surface  vary depending on their speaking style. Note that our neural framework is able to accurately co-register them and compute the 4D mean surface.